%% file: main.tex
\documentclass[10pt,twocolumn,letterpaper]{article}

\usepackage[pagenumbers]{cvpr}

\input{preamble}
\definecolor{cvprblue}{rgb}{0.21,0.49,0.74}
\usepackage[breaklinks,colorlinks,allcolors=cvprblue]{hyperref}

\def\confName{CVPR}
\def\confYear{2027}

\title{OVOK: A Recursive Self-Improving Agentic Framework for Video-Based Articulated Robot State Estimation}

\author{
Cheng Dai$^{1*\ddag}$ \quad
Renkai Luo$^{1,2*}$ \quad
Guanfeng Tang$^{1}$ \quad
Siming Li$^{1}$ \quad
Hao Zhao$^{1\dag}$ \\
$^{1}$AIR, Tsinghua University \quad $^{2}$Qiuzhen College, Tsinghua University
}

\begin{document}
\maketitle
{
  \renewcommand{\thefootnote}{}
  \footnotetext{%
  \begin{minipage}[t]{\dimexpr\linewidth-2em\relax}
  \textsuperscript{\(\ddag\)}Project lead. \textsuperscript{*}Equal contribution.  \textsuperscript{\dag}Corresponding author.
  \end{minipage}}
}
\begin{strip}
  \centering
  \includegraphics[width=\textwidth]{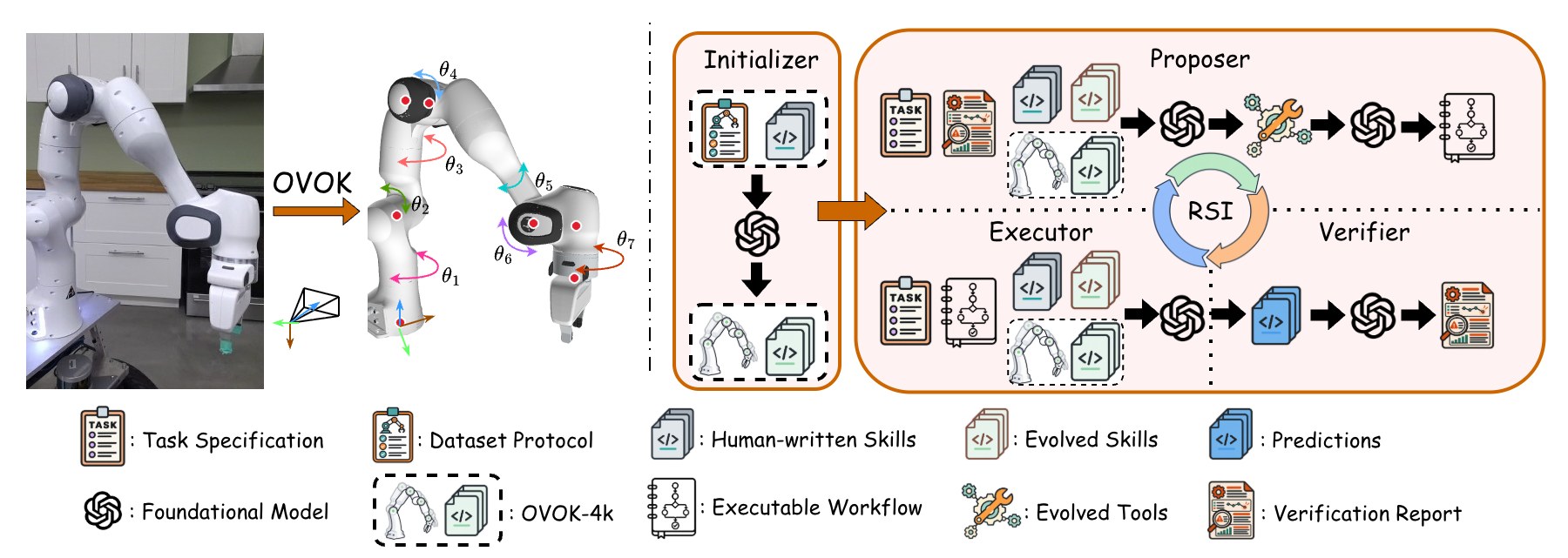}
  \captionof{figure}{Overview of OVOK (One Video to Kinetics).
  Left: an executable workflow maps an RGB video to camera intrinsics, camera-relative robot pose, and joint angles.
  Right: an initializer seeds a robot-specific dataset: OVOK-4k; under recursive self-improvement, a Proposer--Executor--Verifier loop uses feedback to accumulate skills and refine tools and the workflow until a compact solver emerges.}
  \label{fig:overview}
\end{strip}
\input{sec/0_abstract}
\begin{figure}[t]
  \centering
  \includegraphics[width=\columnwidth]{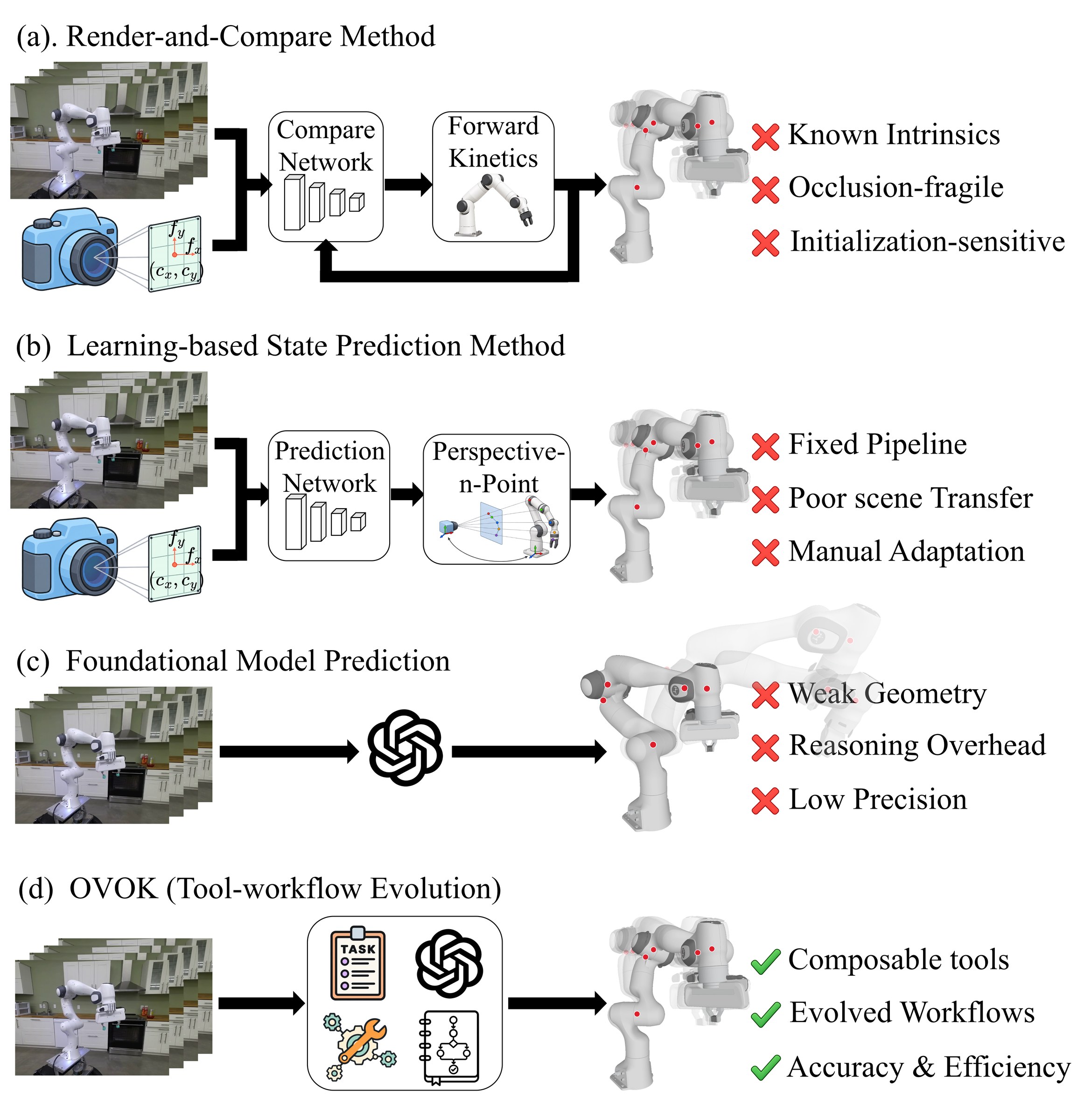}
  \caption{Comparison of paradigms (a--c) against OVOK (d), evolving tools and workflows for robot state estimation.}
  \label{fig:paradigm-comparison}
\end{figure}
\input{sec/1_intro}
\input{sec/2_related}
\input{sec/2b_preliminary}
\input{sec/3_method}
\input{sec/4_experiments}
\input{sec/5_conclusion}
\input{sec/6_llm_usage}
{
    \small
    \bibliographystyle{ieeenat_fullname_citeorder}
    \bibliography{main}
}

\input{sec/A_appendix}

\end{document}

%% file: preamble.tex
\usepackage{cuted}

\usepackage{currfile}

\usepackage{caption}
\usepackage{array}
\usepackage{tabularx}

\usepackage{algorithm}
\usepackage{algpseudocode}
\usepackage{listings}
\usepackage{xcolor}
\lstdefinestyle{ovokprompt}{
  basicstyle=\footnotesize\ttfamily\color{black!88},
  backgroundcolor=\color{black!4},
  breaklines=true,
  breakatwhitespace=false,
  columns=fullflexible,
  keepspaces=true,
  showstringspaces=false,
  frame=single,
  rulecolor=\color{black!28},
  framesep=6pt,
  xleftmargin=2pt,
  xrightmargin=2pt,
  aboveskip=6pt,
  belowskip=4pt,
  literate=
    *{→}{{$\rightarrow$}}1
     {×}{{$\times$}}1
     {≥}{{$\geq$}}1
     {≤}{{$\leq$}}1
     {←}{{$\leftarrow$}}1,
}
\DeclareMathOperator{\FK}{FK}

\newcommand{\vmid}[1]{\raisebox{-1.15ex}[0pt][0pt]{#1}}

\newcommand{\pmstd}[2]{%
  \begin{tabular}[c]{@{}c@{}}#1\\[-0.38ex]{\scriptsize($\pm$#2)}\end{tabular}%
}

%% file: sec/0_abstract.tex
\begin{abstract}
Recovering articulated robot state from video supports metric spatial sensing for manipulation, 
scene reconstruction, and action planning.
Yet the problem is ill-posed: different intrinsics, camera-to-robot poses, and joint angles can explain similar observations.
Prior task-specific regressors typically assume known intrinsics, 
fit one robot through laborious data collection and robot-specific training, and still transfer poorly across scenes.
Foundation models apply to any robot zero-shot, but their estimates have large errors.
We present OVOK (One Video to Kinetics), a recursive self-improving agentic framework for this task.
Under a human-authored protocol, an agentic initializer constructs robot-specific OVOK-4k; a proposer--executor--verifier loop then jointly evolves tools and executable workflows from feedback.
Updated tools and workflows are accepted only when the verifier judges a positive improvement.
The retained workflow is frozen for accurate inference.
Experiments demonstrate that the evolved workflow achieves state-of-the-art results on DREAM-V and RoboKeyGen-V, 
delivering robust and accurate estimates.
\end{abstract}

%% file: sec/1_intro.tex
\section{Introduction}
\label{sec:intro}

Video-based articulated robot state estimation recovers camera intrinsics, camera-relative robot pose, and joint angles from a RGB sequence, given a known metric robot model.
As a metric interface between raw video and physical embodiment, it underpins vision-based manipulation and closed-loop control~\cite{lee2020dream,lu2023ctrnet,wu2025feepe,huang2025rekep,cai2026internvla,chen2026robodojo},
4D scene reconstruction~\cite{liu2025videoartgs,chen2026ovow,delitzas2026funrec},
and fine-grained motion planning~\cite{shen2025cutamp,yang2025vlmtamp,kumar2025owltamp,shcherba2025mops,kim2026cocotamp,huang2026kinder,merler2026agenticgenplan,mendez2025llmtamp,yan2026vlmconstrain}.

Recovering these quantities from RGB remains challenging: even with a known metric robot model, unknown intrinsics and camera-to-robot pose couple projection with articulation~\cite{ban2024horopose,zhu2025monose}, 
while self-occlusion~\cite{tian2023sgtapose}, partial visibility~\cite{goswami2025robopepp,wu2025feepe}, a small image footprint and imaging degradations (e.g., lighting variation and blur) leave only weak geometric evidence~\cite{tian2024robokeygen}.

As illustrated in Fig.~\ref{fig:paradigm-comparison}, prior work mainly follows two task-specific paradigms.
Render-and-compare refinement~\cite{labbe2021robopose,wen2024foundationpose,li2018deepim} is accurate when well initialized, but typically needs known intrinsics and remains fragile to occlusion.
Learning-based state predictors~\cite{tian2024robokeygen,shao2026ropeflow,goswami2025robopepp,ban2024horopose,liu2026robotag} estimate the state, yet depend on robot-specific training, hand-designed pipeline, and limited cross-scene transfer~\cite{tian2024robokeygen}.

Beyond these paradigms, recent multimodal foundation models can be prompted to output articulated robot state directly from RGB~\cite{gemini2025robotics15,openai2026gpt56sol,openai2026gpt6astra,anthropic2026fable51,mazur2026agentstar}.
Yet the estimates remain metrically imprecise~\cite{liang2025trainingfree}, and repeated visual reasoning incurs high overhead.
As illustrated in Fig.~\ref{fig:paradigm-comparison}, zero-shot localization errors and geometric inconsistencies expose a gap between semantic understanding and metric estimation~\cite{han2025tiger}.

Bridging this gap calls for more than a stronger zero-shot predictor.
Recursive self-improvement (RSI) aims to let an autonomous agent acquire knowledge from experience and progressively improve its solver~\cite{zhang2025dgm,xia2026rrsi,robeyns2025sica}: given task feedback, an MLLM~\cite{brown2020language} proposes edits, evaluates them, and retains only those that improve accuracy or cost.
In geometric perception, however, most RSI systems rewrite orchestration or agent code~\cite{zelikman2023stop,hu2025adas,zhang2025aflow,zhang2025dgm,robeyns2025sica,zhang2026selfharness,lin2026ahe,lee2026metaharness,shi2026pluginrsi,wu2026modularrsi,xia2026rrsi,xu2026harness2,karten2026continual,jiang2025aide,srikanth2026aide2,novikov2025alphaevolve,zhu2026rsiaagent,hu2026coap}; they rarely redesign the perception tools that yield metric measurements.

We therefore introduce \textbf{OVOK} (One Video to Kinetics), a task-specific RSI framework that evolves an executable perception-and-geometry workflow~\cite{han2025tiger,chen2025gca,lopopolo2026harness,rajasekaran2026harness},
which estimates the full articulated robot state: camera intrinsics, camera-to-robot pose, and joint angles, from a video alone under a known URDF.
An agentic \emph{initializer} builds OVOK-4k, a synthetic dataset for tools development and workflow evaluation.
During online evolution, a Proposer--Executor--Verifier loop accumulates skills and revises both tools and workflows from feedback~\cite{ye2026mce,zhang2026ace,zhang2025dgm,agrawal2026gepa,zhang2025rlm,zhang2026mgh}.
The retained workflow is then frozen for inference.

Concretely, OVOK targets the failure modes in Fig.~\ref{fig:paradigm-comparison}.
First, visual models are invoked as tools that provide coarse estimates, mitigating occlusion fragility and initialization sensitivity.
Second, OVOK evolves perception and geometry tools and workflows to remove known-intrinsics assumptions, fixed pipelines, and weak scene transfer in prior methods.
Third, from a light human seed: an initialization protocol, an initial skill--tool--workflow triple, and an evolution regularization and schedule, OVOK runs long-horizon self-discovery and revision, retaining a compact workflow with higher metric precision and robustness.

In summary, this work makes four contributions:
\begin{itemize}
    \item \textbf{RSI framework:} We introduce OVOK, a recursive self-improving framework that jointly evolves perception and geometry tools with their executable workflows to recover articulated robot state from uncalibrated RGB video.
    \item \textbf{Closed loop via OVOK-4k:} A human-authored initialization protocol lets the initializer build OVOK-4k, a video set with ground-truth states suppling feedback for tool adaptation, workflow selection, and verification.
    \item \textbf{Light seeds for long-horizon evolution:} With a skill--tool--workflow seed and an evolution regularization and schedule, OVOK sustains long-horizon improvement.
    \item \textbf{Frozen transferable solver:} The evovled workflow are frozen when inference and achieve state-of-the-art accuracy and robustness on DREAM-V and RoboKeyGen-V.
\end{itemize}

%% file: sec/2_related.tex
\section{Related Work}
\label{sec:related}

\subsection{Articulated Robot State Estimation}
\label{sec:related-pose}

When joint angles are known, DREAM~\cite{lee2020dream}, CtRNet~\cite{lu2023ctrnet}, and CtRNet-X~\cite{lu2025ctrnetx} recover camera-to-robot pose using visual correspondences and kinematic geometry.
SGTA-Pose~\cite{tian2023sgtapose} incorporates temporal cues, while MonoSE(3)-Diffusion~\cite{zhu2025monose} refines pose through conditional denoising.

Estimating articulation together with camera-relative pose removes this joint-state requirement.
RoboPose~\cite{labbe2021robopose} uses iterative render-and-compare refinement, while HoRoPose~\cite{ban2024horopose} combines state regression with learned depth estimation.
RoboKeyGen~\cite{tian2024robokeygen} and RoPE-Flow~\cite{shao2026ropeflow} generate 3D keypoints through diffusion and graph-conditioned flow matching, respectively.
RoboPEPP~\cite{goswami2025robopepp} uses joint-masking pretraining for joint-angle and keypoint prediction, followed by geometric pose recovery.
RoboTAG~\cite{liu2026robotag} aligns 2D and 3D branches through a topological graph, reducing dependence on labeled data with consistency supervision.

Depth-based approaches exploit additional geometric observations.
FEEPE~\cite{wu2025feepe} combines measured depth and kinematic motion priors for end-effector pose refinement, while SPDH~\cite{simoni2022spdh} estimates 3D joint locations from depth.
Their reliance on depth measurements limits use where depth sensing is unavailable.

Estimators listed above generally assume known camera intrinsics~\cite{goswami2025robopepp,shao2026ropeflow}, limiting direct use of uncalibrated video.
More broadly, manually designed perception modules and solver pipelines require further engineering to address new observation conditions and failure modes.

\subsection{General Recursive Self-Improvement}
\label{sec:related-rsi}

Alongside approaches that update model parameters through training~\cite{ornith2026selfimprove,zelikman2022star,yuan2024selfrewarding,guo2025deepseekr1,zhao2025absolutezero,huang2025rzero,qi2024webrl}, recent work improves agents by optimizing the harness around a foundation model~\cite{lilianweng2026harness}.
Execution feedback guides harness revisions, and the revised agent produces experience for the next round~\cite{zelikman2023stop,zhang2026selfharness,lin2026ahe,robeyns2025sica}.
Knowledge and executable components retained across iterations then support further improvement~\cite{luo2026physicalrsi,zhao2026memento3}.

These approaches can be organized by what they revise.
One line edits what the model reads, updating prompts and memory from execution outcomes without changing model parameters~\cite{agrawal2026gepa,ye2026mce,yuksekgonul2025textgrad,suzgun2026cheatsheet,zhang2026ace}.
A second line edits what the model can execute: skills accumulate as reusable procedures, and tools provide additional computation~\cite{wang2023voyager,yang2026skillcompiler,li2026roboharness,chen2026showharness,glantz2026seeing,zhao2026memento3}.
A third line edits how components are composed, searching over workflows and agent implementations~\cite{hu2025adas,zhang2025aflow,lee2026metaharness,shi2026pluginrsi,wu2026modularrsi,xia2026rrsi},
which relies on informative feedback and reliable candidate evaluation to retain useful revisions~\cite{xu2026harness2}.

\subsection{Self-Improvement for Geometric Tasks}
\label{sec:related-embodied}

Recent foundation models (e.g., GPT-6-Astra) show strong 3D spatial understanding and measurement ability~\cite{openai2026gpt6astra,menon2026robocurve,anthropic2026opus55,deepseekai2026v4,you2026frontierphysics}.
Yet geometric tasks remain difficult: camera-relative pose, articulated joint angles, and metric distance must be recovered from video under tight latency, and current models yield only coarse estimates rather than precise ones.
Two limitations underlie this gap.
First, models lack explicit metric spatial reasoning grounded in object geometric priors such as a URDF or mesh~\cite{chen2024spatialvlm,cheng2024spatialrgpt,fu2024blink,han2025tiger,chen2025gca}.
Second, they provide no computational tools for measurement and no interface through which such tools can be invoked~\cite{glantz2026seeing}.

Two families of methods address these limitations in turn.
The first strengthens spatial reasoning, refining an estimate through repeated inspection, hypothesis, and revision against the observation~\cite{mazur2026agentstar,chen2025gca,han2025tiger}; accuracy improves, at the cost of inference time.
The second supplies measurement tools, such as depth, calibration, and refiners, and orchestrates them at inference~\cite{li2026roboharness,glantz2026seeing,chen2026showharness,labbe2021robopose,wen2024foundationpose,hu2026coap,wang2026regenharness}.
Their predictions are typically more stable and accurate, yet orchestration is often hand-designed, and few offer an efficient feedback-driven mechanism for revising tools and composition. This is the gap OVOK addresses.

%% file: sec/2b_preliminary.tex
\section{Preliminary}
\label{sec:prelim}

\paragraph{State and Observation}
Let \(\mathcal{M}\) be a known kinematic model of a robot: a tree of links, \(n\) revolute joints with limits \(\mathcal{Q}\subset\mathbb{R}^{n}\), and a body frame on each link. A configuration \(q\in\mathcal{Q}\) determines, by forward kinematics, a rigid pose \(\FK_{j}(q)\in\mathrm{SE}(3)\) of every link \(j\) in the robot-base frame.

An RGB camera with intrinsics \(K\) records a video \(I_{1:N}=(I_t)_{t=1}^{N}\), where \(I_t\in\mathbb{R}^{H\times W\times 3}\).
At frame \(t\), the robot state is \((\mathbf{T}_{c\leftarrow b},q_t)\), where \(\mathbf{T}_{c\leftarrow b}\in\mathrm{SE}(3)\) maps the robot-base frame to the camera frame and \(q_t\) specifies the joint configuration.
For kinematic keypoints \(\mathcal{J}\) defined at selected link origins, the image projections are
\begin{equation}
\label{eq:kine-proj}
  \mathbf{p}_{t,j}
  =
  \pi\bigl(K,\,\mathbf{T}_{c\leftarrow b},\,\FK_{j}(q_{t})\bigr)
  \in\mathbb{R}^{2},
  \qquad j\in\mathcal{J}.
\end{equation}
Conditioned on \(K\) and \(q_t\), forward kinematics supplies 3D keypoints whose image correspondences permit recovery through PnP~\cite{lepetit2009epnp,labbe2021robopose,goswami2025robopepp} under non-degenerate geometry.

\paragraph{Task Definition}
Given \(I_{1:N}\) and \(\mathcal{M}\), we estimate camera intrinsics \(\hat K\), the camera-relative pose \(\hat{\mathbf{T}}_{c\leftarrow b}\), joint angles \(\hat q_t\in\mathcal{Q}\) per frame.
We instantiate this on Franka Emika Panda, taking \(\mathcal{J}\) as the seven link-origin markers.

\paragraph{Formulation Rationale}
Prior work typically takes a single RGB image and predicts 2D/3D keypoints together with joint angles~\cite{labbe2021robopose,ban2024horopose,tian2024robokeygen,goswami2025robopepp,shao2026ropeflow,liu2026robotag}.
With known intrinsics, image--model correspondences then admit metric recovery of camera-relative pose through a PnP solver~\cite{lepetit2009epnp}.
We instead estimate \(K\), \(\mathbf{T}_{c\leftarrow b}\), and \(q_t\) jointly from a video under a known metric model \(\mathcal{M}\), for two reasons.
First, treating \(K\) as given usually presupposes offline camera calibration, laborious in practice and a barrier to uncalibrated recordings.
Second, video is the native observation in deployment: temporal coherence regularizes frames degraded by self-occlusion or semantic ambiguity, while the metric robot model supplies absolute scale, leaving the estimation less severely underdetermined than a single uncalibrated image.

%% file: sec/3_method.tex
\section{Methodology}
\label{sec:method}

\begin{figure}[t]
  \centering
  \includegraphics[width=\columnwidth]{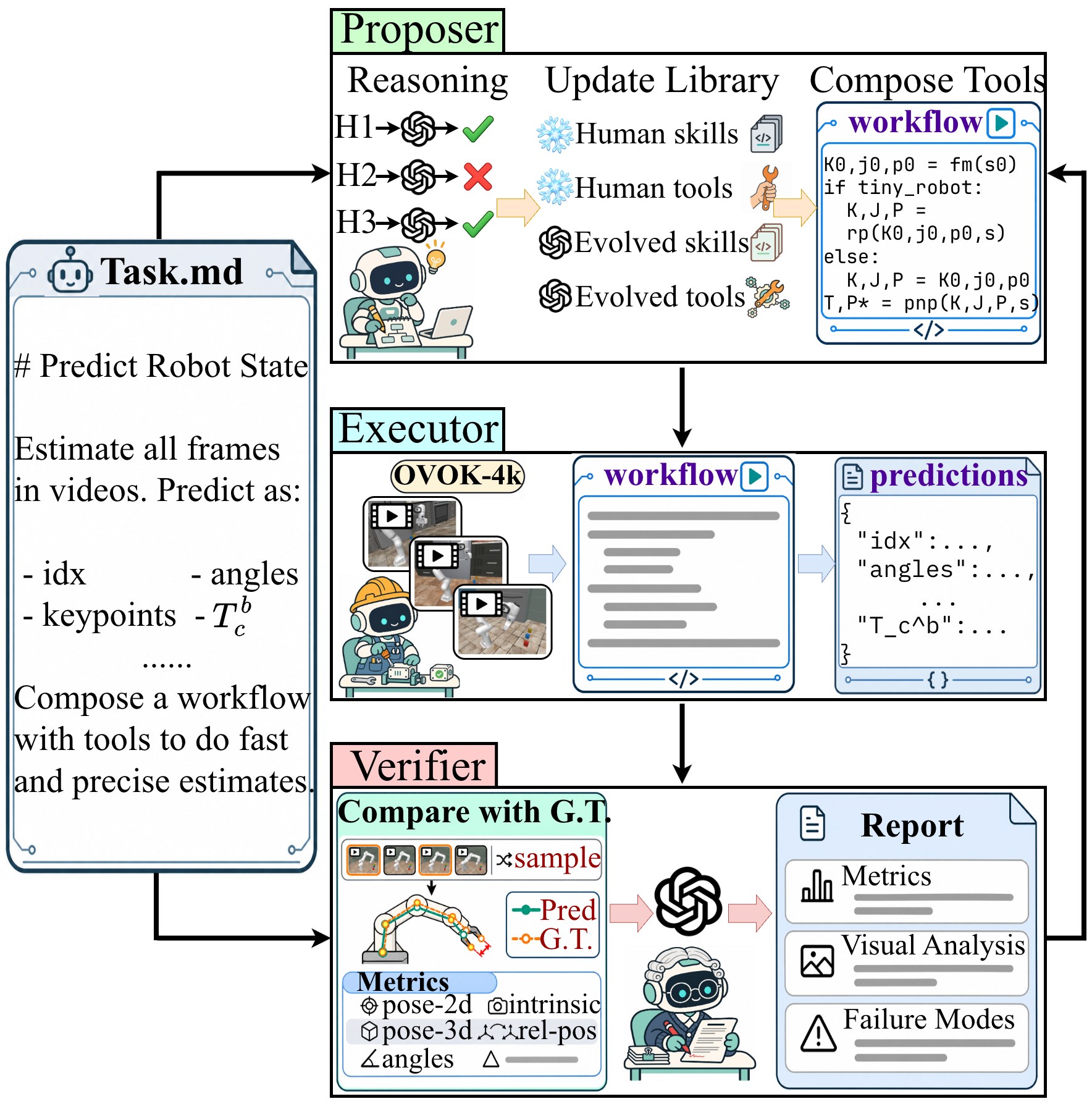}
  \caption{The OVOK loop. A proposer issues hypotheses, revises tools on training data, and composes a frozen workflow. An executor runs it on validation video, and a verifier turns the resulting metrics and failure analysis into edits for the next iteration.}
  \label{fig:loop}
\end{figure}

\subsection{Overview}
\label{sec:overview}

OVOK is a framework that evolves a workflow for articulated robot state estimation.
Its three stages (initialization, evolution, inference) and four roles (initializer, proposer, executor, verifier) are organized around a foundation model.

Initialization establishes the data foundation. Under a human-authored protocol (see Appendix~\ref{app:data}), the initializer
assembles the simulation and rendering pipeline and produces OVOK-4k: 3{,}600 training clips for
developing tools and 400 validation clips for scoring workflows.
Evolution runs the loop of Fig.~\ref{fig:loop}. A proposer issues hypotheses, updates skills, revises tools on the training split,
and composes the workflow; an executor runs the frozen workflow on the validation set; a verifier
audits the outcome, converting measured error and diagnosed failure modes into report that
steers the next revision.
Inference deploys the evolved workflow on real videos, returning estimates efficiently.

\subsection{Evolution Space}
\label{sec:artifacts}

Following Sec.~\ref{sec:overview}, OVOK evolves over three artifacts: skills \(\mathcal{S}\), tools \(\mathcal{U}\), and an executable workflow \(\mathcal{W}\).
Skills \(s\in\mathcal{S}\) record task knowledge~\cite{zhang2026ace,gao2025maps,yuksekgonul2025textgrad,lu2026aspire}:
related works, accumulated hypotheses and open issues.
Tools \(u\in\mathcal{U}\) are callable computations~\cite{hu2025adas,yang2026skillcompiler}.
The workflow \(\mathcal{W}\) binds tools into a program~\cite{zhang2025rlm,zhang2026harnesses,shi2026pluginrsi}.

\paragraph{Evolution Policy} 
During evolution we impose one structural prior: \textit{articulated state estimation admits a modular decomposition into subproblems.} 
Modular decomposition matches how the task is structured and how it must be solved in practice.
Current methods generally use a modular, serial pipeline: keypoint localization and camera-relative pose recovery recur as separable subproblems across estimators~\cite{ban2024horopose,tian2023sgtapose,tian2024robokeygen,shao2026ropeflow,goswami2025robopepp,lepetit2009epnp,lee2020dream,lu2023ctrnet,lu2025ctrnetx,labbe2021robopose,wen2024foundationpose,li2018deepim,liu2026robotag}.
Deployment, however, requires adaptive depth: well-observed clips need only a short route to accurate estimates, while occluded or ambiguous ones need heavier inference~\cite{hu2025adas,zhang2025aflow,wu2026modularrsi}.
Composing those modules into a branchable workflow therefore covers more corner cases than one serial pipeline, attributes failures to the responsible unit so fixes can be reused, and keeps each revision far cheaper than retraining an end-to-end model.
Furthermore, the modular design is efficient and extensible, and adapts to new observations through incremental updates to individual modules.

\paragraph{Skills}
A skill states what a subtask is, when it applies, and what has been learned about it; 
it is the unit through which OVOK carries knowledge across iterations~\cite{wang2023voyager,zhang2026ace}.

Skills come in two kinds. The human-authored seed lets the proposer understand the task before it has any evidence: problem definitions are written in compact mathematical form, and prior work is summarized in plain, structural language. 
The seed stays frozen for the whole run, which keeps the objective from drifting as skills and tools evolve~\cite{xia2026rrsi}.

Evolved skills are written from the verification report and carry two things only: 
validated hypotheses, 
and issues that need solving next~\cite{zhang2026selfharness,xu2026harness2,karten2026continual,lin2026ahe,ye2026mce,gao2025maps,yuksekgonul2025textgrad,lilianweng2026harness}.

\paragraph{Tools.}
Tools fall into two categories. 
Perception tools achieve robust and real-time pattern recognition on the scenes, especially under limited observations.
Computation tools complementarily perform exact geometric computation by deterministic means~\cite{han2025tiger,chen2025gca,glantz2026seeing,li2026roboharness}. 

Initialization ships a frozen pool of backbones (e.g., Depth Anything 3~\cite{lin2025da3}, DINOv3~\cite{simeoni2025dinov3}, and V-JEPA~\cite{bardes2024vjepa}) 
together with a set of mutable tools covering the operations a workflow is likely to need~\cite{han2025tiger,glantz2026seeing}. 
During evolution, tools are revised for the proposed hypotheses and accumulated in \(\mathcal{U}\)~\cite{shi2026pluginrsi}, 
in which task-specific adaptation is further required for perception tools. 
OVOK allows fine-tuning adapters on frozen backbones and training small task-specific models on the training split, 
which takes few hours~\cite{bousmalis2023robocat,kulikov2026autodata}.

\paragraph{Workflow.}
In OVOK, a workflow is an executable Python program~\cite{hu2025adas,zhang2025aflow,lee2026metaharness,zhang2025rlm},
containing conditional branches, data dependencies, and tool bindings. It is composed after the tools are revised 
and stays frozen during execution.

\subsection{Initialization}
\label{sec:initer}

Initialization builds the data foundation. Following a human-authored protocol, the initializer assembles the simulation and rendering pipeline and produces OVOK-4k: 
4{,}000 paired clips of RGB video and simulator state for a Franka Panda in RoboCasa kitchens~\cite{nasiriany2026robocasa365,lee2020dream,ban2024horopose,tian2024robokeygen,cheng2026gptpolicy}, 

OVOK-4k splits into 3{,}600 training clips $\mathcal{D}_{\mathrm{tr}}$ for developing tools and 400 validation clips $\mathcal{D}_{\mathrm{v}}$ 
for scoring workflows, with disjoint room layouts. 
Each clip is admitted only after independently quantitative checks (see Appendix~\ref{app:data}).

\subsection{Evolution}
\label{sec:orchestration}

\paragraph{Evolution Objectives.}
For task \(\tau\) (Sec. \ref{sec:prelim}), OVOK targets:
\begin{equation}
(\mathcal{U}^{\star},\mathcal{W}^{\star})
=\underset{\mathcal{U},\mathcal{W}}{\arg\max}\;
J_{\tau}(\mathcal{U},\mathcal{W};\mathcal{D}_{\mathrm{v}}),
\label{eq:selection-objective}
\end{equation}
where \(J_{\tau}\) measures accuracy and efficiency under a fixed evaluation protocol~\cite{srikanth2026aide2,xia2026rrsi,ye2026mce,shi2026pluginrsi} (see Appendix.\ref{app:proposer}).

\paragraph{Evolution Progress.}
To optimize (Eq.~\ref{eq:selection-objective}), OVOK runs a proposer--executor--verifier cycle.
The proposer reads skills, tools, and the previous workflow together with the verification report,
and raises \(M\) hypotheses \(\{\delta_m\}_{m=1}^{M}\) that target distinct aspects of the solver.
It starts from a human-initialized \((\mathcal{U}_0,\mathcal{S}_0,\mathcal{W}_0)\) with empty report \(r_0=\varnothing\)~\cite{zhu2026rsiaagent} (see Alg.~\ref{alg:seed-workflow}).
Given the hypotheses, the proposer revises tools on \(\mathcal{D}_{\mathrm{tr}}\), yielding \(\widetilde{\mathcal{U}}_i\), then composes a workflow.
Candidates that fail a cheap dry-run probe are rejected before full validation~\cite{agrawal2026gepa,zhang2025aflow,moll2026grasp,hu2025adas,ye2026mce,xu2026harness2}. The executor then runs the composed workflow on the validation split. 

The verifier inspects each run in two passes.
Quantitatively, it measures errors in 2D/3D pose, camera-to-base pose, intrinsics, joint angles, and related quantities, while also recording throughput.
Qualitatively, it samples high-error clips, diagnoses their failure patterns, and aggregates them into failure modes.
Both analyses are written into a verification report that steers the next round.
A revision is retained only when \(J_i>J^{\star}\), i.e., when it improves either precision or efficiency (see Appendix~\ref{app:proposer}).

\paragraph{Regularization and Schedule.}
Unconstrained LLM feedback can mis-attribute causes, pursue biased local gains, and take uncontrolled edit steps~\cite{yuksekgonul2025textgrad,huang2026textresnet,xia2026rrsi,zhang2025dgm}.
We therefore inject two human controls into the RSI loop.

First, attribution is confined to eight fixed failure modes: inaccurate intrinsics estimation, severe \(\mathbf{T}_{c\leftarrow b}\) bias, global joint-angle bias, difficult base-orientation estimation (J1), difficult end-effector estimation (J7), occlusion/partial identifiability, lighting, and low computational efficiency.

Second, we advise a staged edit planning: first prioritize locking camera intrinsics \(K\), which can be readily estimated from environmental background texture~\cite{veicht2024geocalib,tirado2025anycalib} and is a prerequisite for metric recovery; once \(K\) is judged adequate, focus on \(\mathbf{T}_{c\leftarrow b}\) and the joint estimators, which dominate final pose accuracy; after these geometric foundations meet the acceptance criteria, 
subsequent edits may improve efficiency while accuracy does not regress a lot (see Appendix~\ref{app:proposer}).

\subsection{Inference}
\label{sec:inference}

In real use, OVOK freezes the evolved configuration $(\mathcal{U}^{\star},\mathcal{S}^{\star},\mathcal{W}^{\star})$.
For an observed video $X$, the system executes the composed workflow directly:
\begin{equation}
\widehat{Y}=\textsc{Execute}(X;\mathcal{U}^{\star},\mathcal{W}^{\star}).
\end{equation}
A single pass returns, for every frame, camera-relative pose, joint angles, intrinsics, and 2D/3D keypoints.

%% file: sec/4_experiments.tex
\section{Experiments}
\label{sec:exp}

\subsection{Implementation Details}
\label{sec:exp-impl}

\begin{figure}[!t]
  \begin{algorithm}[H]
    \caption{Initialized workflow \(\mathcal{W}_0\)}
    \label{alg:seed-workflow}
    \footnotesize
    \begin{algorithmic}[1]
    \Require video \(I_{1:N}=(I_t)_{t=1}^{N}\), robot model \(\mathcal{M}\), tools \(\mathcal{U}_0\)
    \Ensure \(\hat K\), \(\hat{\mathbf{T}}_{c\leftarrow b}\), \(\{\hat q_t\}_{t=1}^{N}\)
    \State \(\hat K\gets\textsc{AnyCalib}(I_1)\)~\cite{tirado2025anycalib}
    \For{\(t=1,\ldots,N\)}
        \State \((\hat{\mathbf{T}}_{c\leftarrow b,t},\hat q_t)\gets\textsc{HoRoPose}(I_t,\hat K;\mathcal{M})\)~\cite{ban2024horopose}
    \EndFor
    \State \(\hat{\mathbf{T}}_{c\leftarrow b}\gets\frac{1}{N}\sum_{t=1}^{N}\hat{\mathbf{T}}_{c\leftarrow b,t}\)
    \State \Return \(\hat K\), \(\hat{\mathbf{T}}_{c\leftarrow b}\), \(\{\hat q_t\}_{t=1}^{N}\)
    \end{algorithmic}
  \end{algorithm}
  \vspace{0.35em}
  \begin{algorithm}[H]
    \caption{Proposer--Executor--Verifier cycle in OVOK}
    \label{alg:online-rsi}
    \footnotesize
    \begin{algorithmic}[1]
    \Require task \(\tau\), \((\mathcal U_0,\mathcal S_0,\mathcal W_0)\), \(\mathcal D_{\mathrm{tr}}\), \(\mathcal D_{\mathrm{v}}=(X_{\mathrm{v}},Y_{\mathrm{v}})\), iterations \(N\)
    \Ensure confirmed \((\mathcal U^\star,\mathcal S^\star,\mathcal W^\star)\)
    \State \((\mathcal U^\star,\mathcal S^\star,\mathcal W^\star)\gets(\mathcal U_0,\mathcal S_0,\mathcal W_0)\)
    \State \(J^\star\gets 0,\quad r_0\gets\varnothing,\quad r^\star\gets\varnothing\)
    \For{\(i=1,\ldots,N\)}
        \State \(\{\delta_m\}_{m=1}^{M}\gets\textsc{Reason}(\tau,\mathcal S_{i-1},\mathcal U_{i-1},\mathcal W_{i-1},r_{i-1},M)\)
        \State \(\widetilde{\mathcal U}_i\gets\textsc{Revise}(\tau,\{\delta_m\}_{m=1}^{M},\mathcal U_{i-1},\mathcal D_{\mathrm{tr}})\)
        \State \(\widetilde{\mathcal W}_i\gets\textsc{Compose}(\tau,\{\delta_m\}_{m=1}^{M},\widetilde{\mathcal U}_i)\)
        \State \((\widehat Y_i,\rho_i)\gets\textsc{Execute}(X_{\mathrm{v}};\widetilde{\mathcal U}_i,\widetilde{\mathcal W}_i)\)
        \State \((J_i,r_i)\gets\textsc{Verify}(\tau,\widehat Y_i,Y_{\mathrm{v}},\rho_i)\)
        \If{\(J_i > J^\star\)}
            \State \((\mathcal U_i,\mathcal S_i,\mathcal W_i)\gets(\widetilde{\mathcal U}_i,\textsc{Merge}(\mathcal S_{i-1},\{\delta_m\}_{m=1}^{M}),\widetilde{\mathcal W}_i)\)
            \State \((\mathcal U^\star,\mathcal S^\star,\mathcal W^\star,r^\star,J^\star)\gets(\mathcal U_i,\mathcal S_i,\mathcal W_i,r_i,J_i)\)
        \Else
            \State \((\mathcal U_i,\mathcal S_i,\mathcal W_i)\gets(\mathcal U_{i-1},\mathcal S_{i-1},\mathcal W_{i-1})\)
        \EndIf
    \EndFor
    \State \Return \((\mathcal U^\star,\mathcal S^\star,\mathcal W^\star)\)
    \end{algorithmic}
  \end{algorithm}
\end{figure}

\paragraph{Evaluation Data}  
No established video-level protocol exists for robot pose and joint-angle estimation.
We extract motion clips from DREAM-real~\cite{lee2020dream} and RoboKeyGen-Franka~\cite{tian2024robokeygen}, denoted DREAM-V and RoboKeyGen-V.

DREAM-V contains \(59\) paired clips from camera setups (Azure Kinect, Kinect 360, RealSense, and Orb), with mean duration \(20.5\,\mathrm{s}\).
RoboKeyGen-V contains \(14\) paired clips from two cameras (RealSense and Azure Kinect), with mean duration \(10.3\,\mathrm{s}\).
Both are real-world recordings of a moving Panda, encoded at \(30\,\mathrm{fps}\).
Within each video the camera and robot base remain relatively static.

\paragraph{Experimental Protocol} 
We evaluate three groups of methods under a shared protocol.
Task-specific models are RoboPose~\cite{labbe2021robopose}, HoRoPose~\cite{ban2024horopose}, and RoboPEPP~\cite{goswami2025robopepp}.
Some methods are not included~\cite{tian2024robokeygen,liu2026robotag,shao2026ropeflow,zhu2025monose} due to unavailable public implementation.
Foundation models are Opus 5.5~\cite{anthropic2026opus55}, GPT-6 Astra~\cite{openai2026gpt6astra}, and GPT-6.1 Sol~\cite{openai2026gpt61sol}, all run at the medium reasoning-effort setting.
We also run them inside the OVOK framework.
Standalone foundation-model baselines use \(10\) independent runs; OVOK runs \(15\) RSI iterations and reports the best retained workflow.
Unless otherwise noted, qualitative comparisons use GPT-6.1 Sol as the default OVOK backbone.
All experiments run in a sandbox with 32 cores of an AMD EPYC 7742, one NVIDIA A800 GPU ($80\,\mathrm{GB}$), $128\,\mathrm{GB}$ host memory, and a $1\,\mathrm{TB}$ storage limit.
More details are in Appendix~\ref{app:exp-setup}.

\paragraph{Experimental Metrics}
Comparative experiments focus on pose accuracy, joint-angle accuracy, and efficiency.
We report 2D pose as PCK at \(2\) pixels, 3D pose as ADD~\cite{hinterstoisser2012add} at \(20\,\mathrm{mm}\) over keypoints in \(\mathcal{J}\),
joint-angle error as MAE in degrees, and efficiency as mean FPS under serial per-video processing.
For RSI analysis, we instead measure the accuracy of the underlying factors together with efficiency:
intrinsics \(K\) by MAPE (mean absolute percentage error),
camera-to-base pose \(\mathbf{T}_{c\leftarrow b}\) by MAE in meters,
joint angles by MAE in degrees, and throughput by FPS.

\subsection{Robot State Estimation Results}
\label{sec:exp-results}

\begin{figure}[t]
  \centering
  \includegraphics[width=\columnwidth]{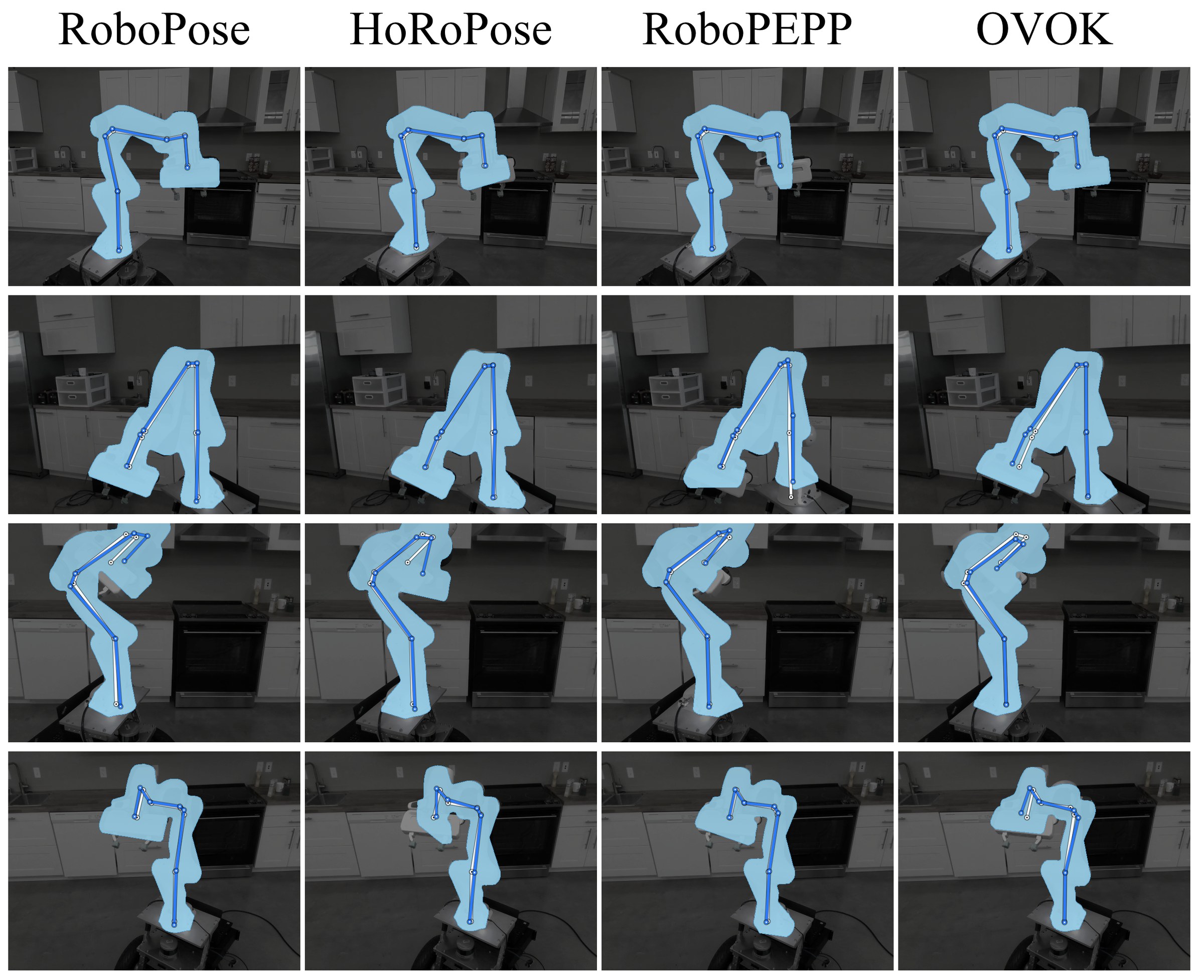}
  \caption{Qualitative comparison on DREAM-V (left to right: RoboPose, HoRoPose, RoboPEPP, OVOK with GPT-6.1 Sol). Blue overlays denote predictions; white denotes ground truth.}
  \label{fig:dream-qual}
\end{figure}

\begin{figure}[t]
  \centering
  \includegraphics[width=\columnwidth]{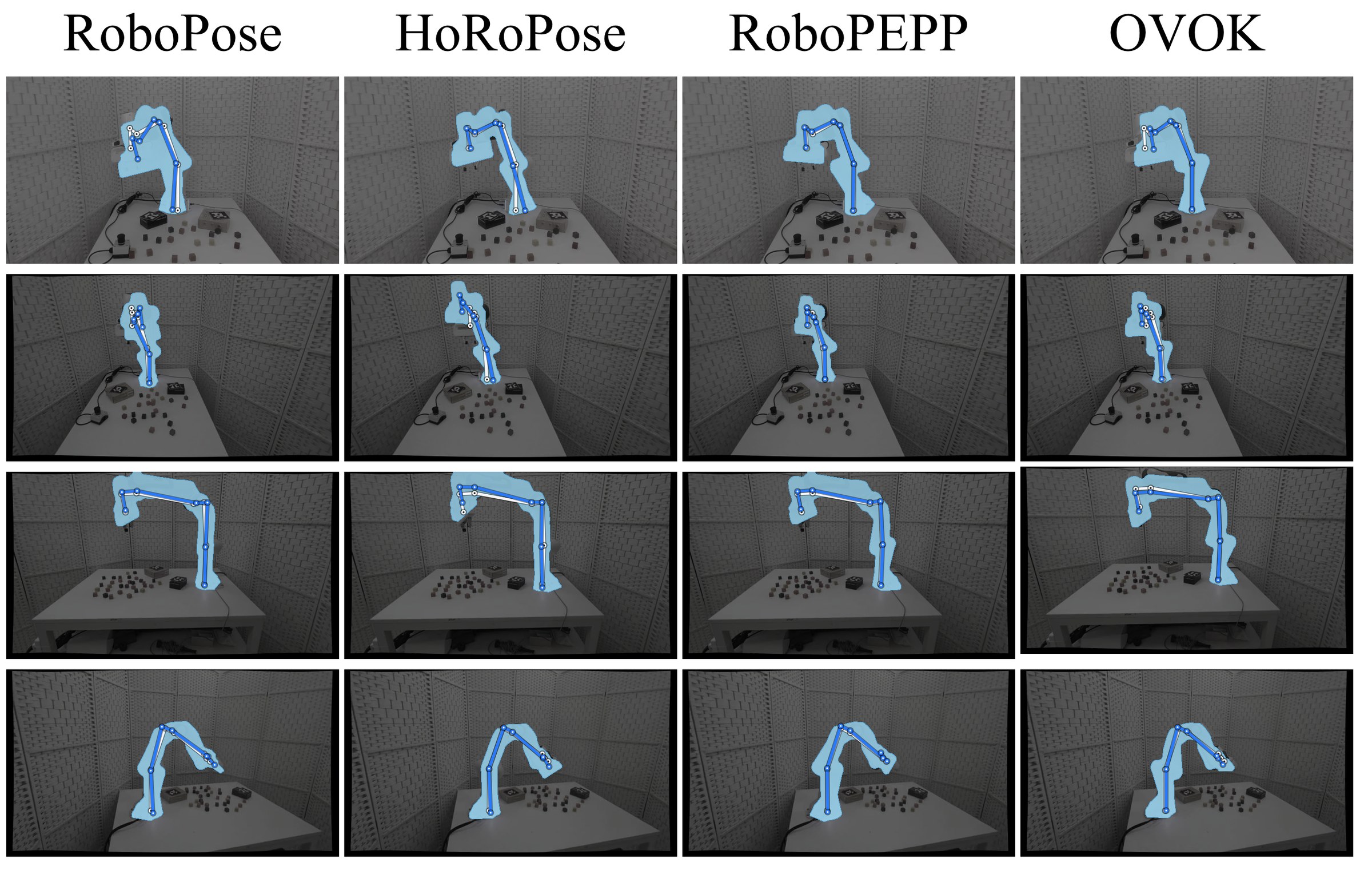}
  \caption{Qualitative comparison on RoboKeyGen-V (left to right: RoboPose, HoRoPose, RoboPEPP, OVOK with GPT-6.1 Sol). Blue overlays denote predictions; white denotes ground truth.}
  \label{fig:rkg-qual}
\end{figure}

Table~\ref{tab:pose} reports 2D/3D pose accuracy and throughput; Table~\ref{tab:joint} reports per-joint angle MAE.
Figures~\ref{fig:dream-qual} and~\ref{fig:rkg-qual} show qualitative overlays for comparison.

\begin{table}[t]
\centering
\footnotesize
\caption{Robot pose estimation on DREAM-V and RoboKeyGen-V.
PCK at \(2\) pixels and ADD at \(20\,\mathrm{mm}\) are success rates.
\(\checkmark\): given at test time.
\(^{*}\): OVOK-solver evolved via foundational model.}
\label{tab:pose}
\setlength{\tabcolsep}{0pt}
\resizebox{\columnwidth}{!}{%
\begin{tabular}{
  l
  @{\hspace{2.2pt}}c
  @{\hspace{2.4pt}}c
  @{\hspace{7.6pt}}c
  @{\hspace{6.8pt}}c
  @{\hspace{6.8pt}}c
  @{\hspace{10.4pt}}c
  @{\hspace{6.8pt}}c
  @{\hspace{6.8pt}}c
}
\toprule
\vmid{Method} & \vmid{Bbox} & \vmid{$K$} &
\multicolumn{3}{c}{DREAM-V} &
\multicolumn{3}{c}{RoboKeyGen-V} \\
\cmidrule(lr){4-6} \cmidrule(lr){7-9}
& & & PCK\,$\uparrow$ & ADD\,$\uparrow$ & FPS\,$\uparrow$ & PCK\,$\uparrow$ & ADD\,$\uparrow$ & FPS\,$\uparrow$ \\
\midrule
\multicolumn{9}{l}{\textit{Task-specific methods}} \\
\midrule
RoboPose &
\(\times\) & \(\checkmark\) & 10.86 & 36.34 & \underline{29.79} & 9.03 & 0.71 & 17.07 \\
HoRoPose &
\(\checkmark\) & \(\checkmark\) & 17.09 & 46.19 & \textbf{30.68} & 12.82 & 1.66 & \underline{21.85} \\
RoboPEPP &
\(\checkmark\) & \(\checkmark\) & 20.60 & 62.78 & 23.43 & \underline{18.63} & 2.04 & 13.06 \\
\midrule
\multicolumn{9}{l}{\textit{Foundation models}} \\
\midrule
Opus-5.5 &
\(\times\) & \(\times\) & \pmstd{3.49}{1.54} & \pmstd{0.00}{0.00} & 19.42 & \pmstd{5.09}{1.17} & \pmstd{0.00}{0.00} & \textbf{27.26} \\
GPT-6-Astra &
\(\times\) & \(\times\) & \pmstd{1.02}{0.46} & \pmstd{0.00}{0.00} & 18.99 & \pmstd{2.74}{1.08} & \pmstd{0.00}{0.00} & 21.45 \\
GPT-6.1-sol &
\(\times\) & \(\times\) & \pmstd{0.62}{0.52} & \pmstd{0.00}{0.00} & 8.53 & \pmstd{1.77}{1.01} & \pmstd{0.00}{0.00} & 3.91 \\
\midrule
\multicolumn{9}{l}{\textit{Foundation models in OVOK}} \\
\midrule
Opus-5.5\(^{*}\) &
\(\times\) & \(\times\) & \underline{28.17} & \underline{73.75} & 2.09 & 17.73 & \underline{4.50} & 1.52 \\
GPT-6-Astra\(^{*}\) &
\(\times\) & \(\times\) & 25.19 & 67.20 & 1.77 & 17.00 & 4.31 & 1.67 \\
GPT-6.1-Sol\(^{*}\) &
\(\times\) & \(\times\) & \textbf{40.32} & \textbf{78.54} & 2.47 & \textbf{23.81} & \textbf{5.88} & 1.93 \\
\bottomrule
\end{tabular}%
}
\end{table}

\begin{table*}[t]
\centering
\footnotesize
\caption{Robot joint-angle estimation on DREAM-V and RoboKeyGen-V.
Entries are per-joint MAE in degrees (lower is better); Avg.\ is the mean over reported joints.
\(\checkmark\): given at test time.
\(^{*}\): OVOK-solver evolved via foundational model.}
\label{tab:joint}
\setlength{\tabcolsep}{0pt}
\renewcommand{\arraystretch}{1.05}
\resizebox{\textwidth}{!}{%
\begin{tabular}{
  l
  @{\hspace{2.2pt}}c
  @{\hspace{2.4pt}}c
  @{\hspace{4.8pt}}c
  *{7}{@{\hspace{3.4pt}}c}
  @{\hspace{7.2pt}}c
  *{7}{@{\hspace{3.4pt}}c}
}
\toprule
\vmid{Method} & \vmid{Bbox} & \vmid{$K$} &
\multicolumn{8}{c}{DREAM-V} &
\multicolumn{8}{c}{RoboKeyGen-V} \\
\cmidrule(lr){4-11} \cmidrule(lr){12-19}
& & &
J1 & J2 & J3 & J4 & J5 & J6 & J7 & Avg.\(\downarrow\) &
J1 & J2 & J3 & J4 & J5 & J6 & J7 & Avg.\(\downarrow\) \\
\midrule
\multicolumn{19}{l}{\textit{Task-specific methods}} \\
\midrule
RoboPose &
\(\times\) & \(\checkmark\) &
\underline{14.21} & 6.92 & 7.79 & \underline{2.67} & 4.40 & 4.19 & 34.49 & 10.67 &
122.09 & 34.70 & 139.26 & 10.69 & 9.67 & 10.83 & 94.29 & 60.22 \\
HoRoPose &
\(\checkmark\) & \(\checkmark\) &
17.31 & \underline{3.75} & 7.24 & 3.42 & 4.99 & 6.09 & 31.03 & 10.55 &
108.34 & 32.62 & 126.76 & \textbf{6.83} & \underline{5.70} & \underline{7.12} & 19.68 & 43.86 \\
RoboPEPP &
\(\checkmark\) & \(\checkmark\) &
66.13 & \textbf{2.84} & 4.24 & \textbf{2.16} & 4.00 & \textbf{2.97} & -- & -- &
119.47 & 34.12 & 133.47 & 13.69 & 10.54 & 9.56 & -- & -- \\
\midrule
\multicolumn{19}{l}{\textit{Foundation models}} \\
\midrule
Opus-5.5 &
\(\times\) & \(\times\) &
\pmstd{39.90}{10.15} & \pmstd{23.71}{13.87} & \pmstd{57.55}{18.48} & \pmstd{25.43}{18.71} & \pmstd{55.41}{28.32} & \pmstd{43.05}{9.15} & \pmstd{53.13}{13.14} & 42.60 &
\pmstd{127.64}{16.85} & \pmstd{39.97}{17.12} & \pmstd{124.41}{13.78} & \pmstd{16.59}{3.90} & \pmstd{43.92}{12.26} & \pmstd{28.76}{10.36} & \pmstd{33.82}{14.76} & 59.30 \\
GPT-6-Astra &
\(\times\) & \(\times\) &
\pmstd{49.08}{13.39} & \pmstd{21.27}{6.22} & \pmstd{45.55}{13.15} & \pmstd{18.21}{4.97} & \pmstd{50.99}{9.10} & \pmstd{47.38}{14.38} & \pmstd{58.51}{11.65} & 41.57 &
\pmstd{125.26}{21.96} & \pmstd{36.13}{9.87} & \pmstd{135.46}{20.35} & \pmstd{19.80}{4.51} & \pmstd{47.03}{13.53} & \pmstd{47.19}{12.37} & \pmstd{42.36}{15.58} & 64.75 \\
GPT-6.1-Sol &
\(\times\) & \(\times\) &
\pmstd{51.37}{9.31} & \pmstd{27.87}{5.52} & \pmstd{60.52}{14.03} & \pmstd{19.66}{5.17} & \pmstd{59.50}{9.26} & \pmstd{46.99}{10.36} & \pmstd{62.48}{15.11} & 46.91 &
\pmstd{110.27}{8.25} & \pmstd{46.07}{13.94} & \pmstd{117.60}{6.68} & \pmstd{22.66}{4.71} & \pmstd{49.46}{10.33} & \pmstd{49.24}{12.26} & \pmstd{48.36}{13.28} & 63.38 \\
\midrule
\multicolumn{19}{l}{\textit{Foundation models in OVOK}} \\
\midrule
Opus-5.5\(^{*}\) &
\(\times\) & \(\times\) &
17.54 & 5.48 & \textbf{3.25} & 3.12 & \underline{3.93} & 4.33 & 14.95 & \underline{7.51} &
\underline{27.71} & \textbf{9.36} & \underline{13.27} & 12.35 & 15.20 & 11.60 & \underline{17.57} & \underline{15.29} \\
GPT-6-Astra\(^{*}\) &
\(\times\) & \(\times\) &
17.87 & 7.16 & 5.59 & 4.73 & 5.25 & 7.00 & \underline{14.18} & 8.83 &
33.46 & 13.45 & 14.15 & 12.62 & 19.39 & 9.95 & 27.03 & 18.58 \\
GPT-6.1-Sol\(^{*}\) &
\(\times\) & \(\times\) &
\textbf{13.43} & 6.26 & \underline{3.82} & 3.54 & \textbf{2.72} & \underline{3.85} & \textbf{8.65} & \textbf{6.04} &
\textbf{20.38} & \underline{9.83} & \textbf{13.08} & \underline{7.38} & \textbf{11.89} & \textbf{5.01} & \textbf{8.73} & \textbf{10.90} \\
\bottomrule
\end{tabular}%
}
\end{table*}

OVOK achieves state-of-the-art pose accuracy and the lowest average joint-angle error among the evaluated methods, without additional inputs.
With GPT-6.1 Sol, it reaches PCK/ADD success rates of \(40.32\%/78.54\%\) on DREAM-V and \(23.81\%/5.88\%\) on RoboKeyGen-V.
These gains are obtained at throughputs of \(2.47\) and \(1.93\)\,FPS, respectively. The relatively higher latency stems from more complex task setting and longer multi-stage inference pipeline.

Standalone foundation-model querying almost collapses under the same setting: ADD success is near zero and average joint-angle MAE remains tens of degrees.
This demonstrates a lack of explicit metric spatial reasoning grounded in articulated robot priors, so direct inference cannot enforce kinematic consistency or recover precise geometry.

Joint-angle errors reveal where visual estimation becomes difficult (Table~\ref{tab:joint}).
J1 depends on subtle base appearance cues, which become less informative as the robot footprint shrinks.
On RoboKeyGen-V, side views and self-occlusion under compact, near-limit arm configurations further challenge DREAM-trained estimators, consistent with large J1--J3 errors when the kinematic chain is poorly observed~\cite{lee2020dream,labbe2021robopose,ban2024horopose}.
OVOK builds on OVOK-4k, which samples diverse viewpoints and scenes~\cite{nasiriany2026robocasa365,tian2024robokeygen}, reducing average joint-angle error to \(6.04^\circ\) on DREAM-V and \(10.90^\circ\) on RoboKeyGen-V.
The improvement is particularly pronounced at J1 and J3 on RoboKeyGen-V, where GPT-6.1 Sol reduces MAE to \(20.38^\circ\) and \(13.08^\circ\).

Pose accuracy decreases substantially on RoboKeyGen-V across all methods (Table~\ref{tab:pose}).
Camera--base distances are larger on RoboKeyGen-V (\(1.29\)--\(1.82\,\mathrm{m}\)) than on DREAM-synthetic (\(0.75\)--\(1.20\,\mathrm{m}\)), the common training set for task-specific estimators~\cite{labbe2021robopose,goswami2025robopepp,ban2024horopose}; the resulting smaller footprint and weaker correspondences make imprecise localization costly under the \(20\,\mathrm{mm}\) ADD and \(2\)-pixel PCK thresholds.
For OVOK the difficulty is stricter: 2D/3D pose accuracy is jointly determined by \(K\), \(\mathbf{T}_{c\leftarrow b}\), and \(q\), and a small bias in any factor is amplified through projection.
Even with an evolved workflow that couples geometric and perceptual tools, high-precision pose remains hard to recover.

The qualitative comparisons illustrate these differences (Figs.~\ref{fig:dream-qual} and~\ref{fig:rkg-qual}).
On DREAM-V, OVOK generally aligns the projected robot structure more closely with the observed arm, including configurations where overlapping links make joints difficult to distinguish.
On RoboKeyGen-V, the small robot footprint makes errors in base alignment and articulated shape more apparent.
OVOK preserves the observed configuration consistently, complementing its improvements in robot pose and joint angle accuracy.

\subsection{Analysis of the RSI Process}
\label{sec:exp-rsi}

\begin{figure}[t]
  \centering
  \includegraphics[width=\columnwidth]{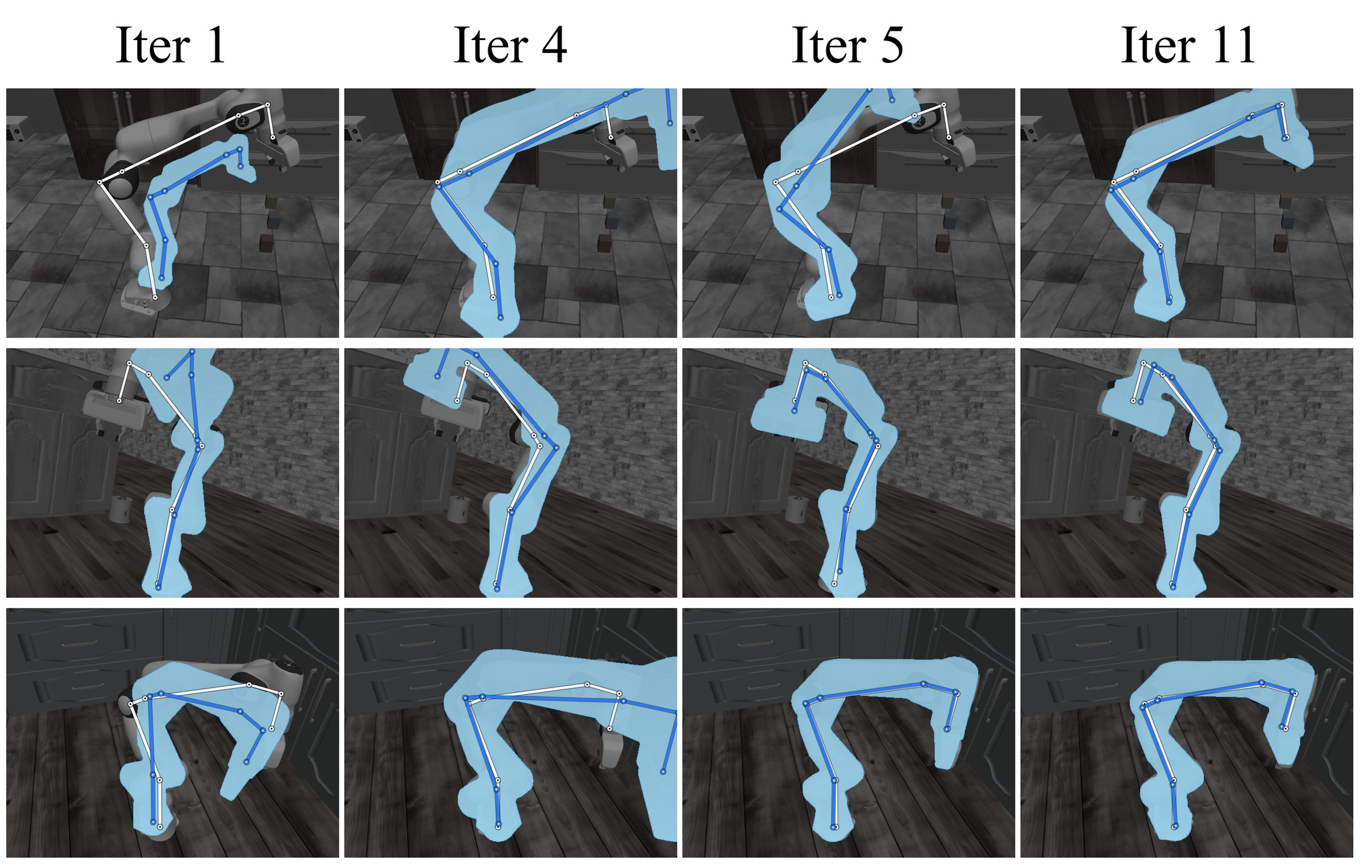}
  \caption{GPT-6.1 Sol$^{*}$ RSI workflows on the OVOK-4k validation set at iterations 1, 4, 5, and 11 (columns) on three clips (rows). Blue overlays denote predictions; white denotes ground truth.}
  \label{fig:rsi-qual}
\end{figure}

\begin{figure*}[t]
  \centering
  \includegraphics[width=\textwidth]{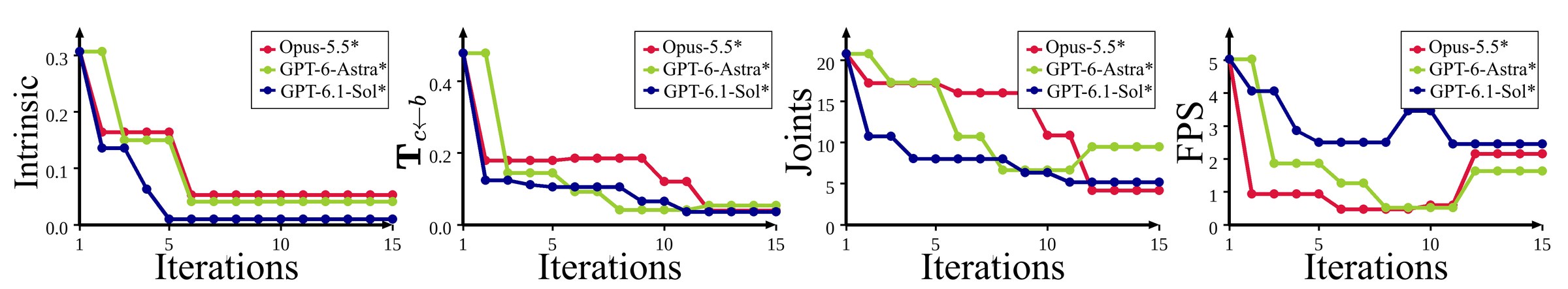}
  \caption{RSI trajectories on the OVOK-4k validation set over 15 iterations for three OVOK backbones (Opus-5.5$^{*}$, GPT-6-Astra$^{*}$, GPT-6.1-Sol$^{*}$).
  Left to right: camera intrinsics MAPE, camera-to-base pose error (m), joint-angle MAE ($^\circ$), and inference FPS.}
  \label{fig:rsi-curves}
\end{figure*}

Fig.~\ref{fig:rsi-curves} plots RSI trajectories of the three OVOK backbones on the OVOK-4k validation set, and Fig.~\ref{fig:rsi-qual} shows four retained GPT-6.1 Sol$^{*}$ workflows on the same split.

Different backbones adopt different search strategies under the same regularization and schedule.
GPT-6.1 Sol$^{*}$ initially treats refining \(K\) and jointly improving \(\mathbf{T}_{c\leftarrow b}\) with joint angles as largely orthogonal subtasks and attacks them in parallel.
In iterations 2--5 it revises the intrinsics tools for rapid \(K\) convergence, while freezing a DINOv3~\cite{simeoni2025dinov3} backbone and training an MLP head to predict sparse-frame joint angles, auxiliary 2D keypoints, and camera-to-base translation/rotation, yielding accurate geometric estimates.
In iterations 9 and 11 it instead freezes the HoRoPose~\cite{ban2024horopose} backbone and fine-tunes JointNet, RotationNet, and KeypointNet with emphasis on J1 and J7, improving both accuracy and efficiency.
GPT-6-Astra$^{*}$, by contrast, concentrates early search almost exclusively on intrinsics: it proposes many \(K\)-centric hypotheses and fine-tunes the AnyCalib~\cite{tirado2025anycalib} decoder for precise calibration, then from iteration~5 shifts to \(\mathbf{T}_{c\leftarrow b}\) and joints via HoRoPose/RoboPose adaptation,
and finally raises FPS by pruning workflow complexity.
Opus-5.5$^{*}$ follows a similar staged path.

Despite these implementation differences, the successful runs share the same macroscopic order: lock camera intrinsics \(K\), which can be readily estimated from environmental background texture~\cite{veicht2024geocalib,tirado2025anycalib}, then refine \(\mathbf{T}_{c\leftarrow b}\) and the joint estimators that dominate final pose accuracy, and only afterwards prune the workflow for efficiency.
As shown in Fig.~\ref{fig:rsi-qual}, early GPT-6.1 Sol$^{*}$ workflows first exhibit large projection error; subsequent revisions gradually tighten the skeleton;
and later iterations refine the estimate toward a precise alignment with the observed arm.

\subsection{Ablation Study}
\label{sec:exp-ablation}

\begin{figure}[t]
  \centering
  \includegraphics[width=\columnwidth]{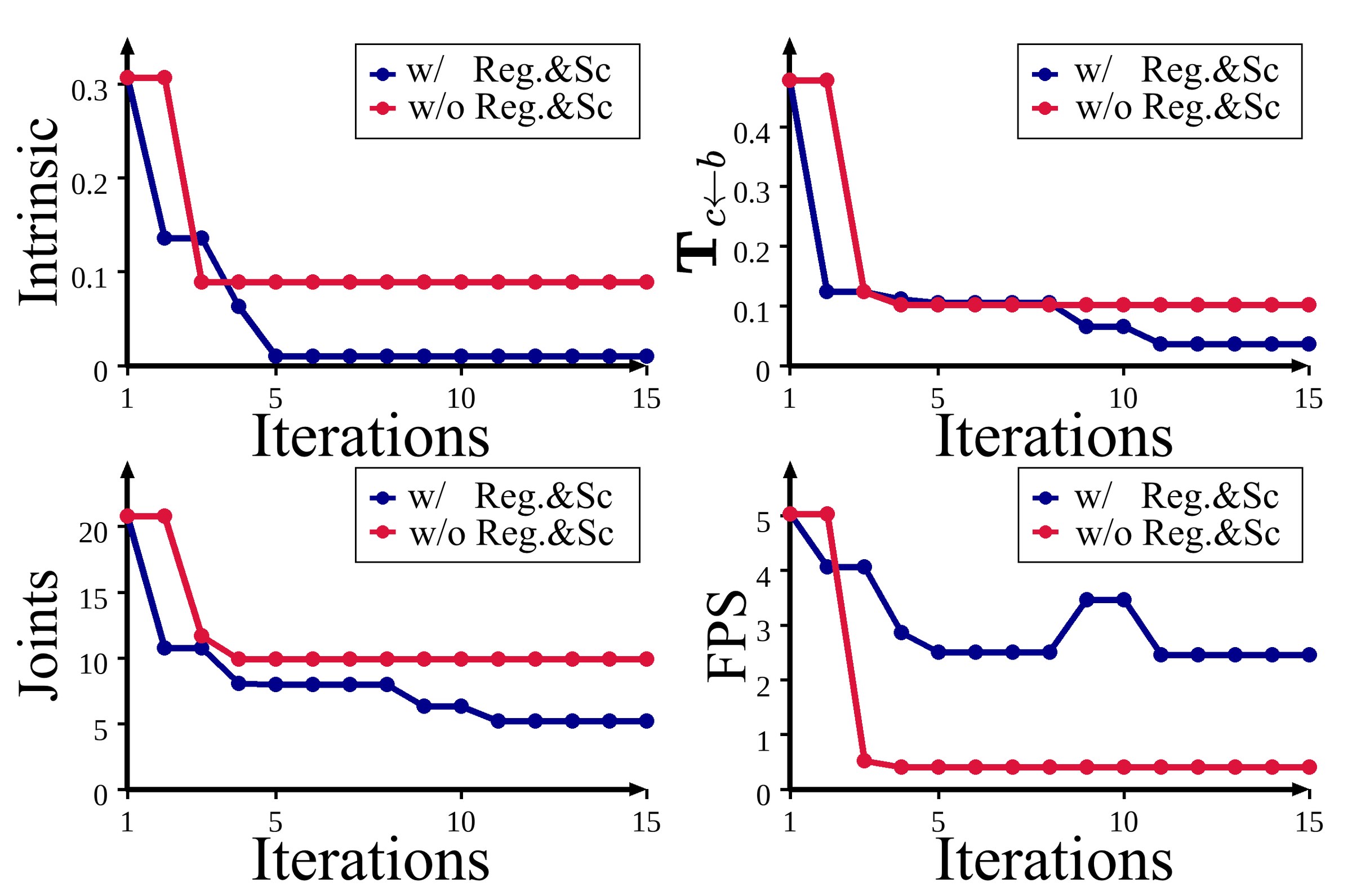}
  \caption{Ablation of human-authored regularization and schedule for GPT-6.1 Sol$^{*}$ on the OVOK-4k validation set.}
  \label{fig:ablation-reg}
\end{figure}

As shown in Fig.~\ref{fig:ablation-reg}, we ablate the human-authored regularization and schedule for GPT-6.1 Sol$^{*}$ on the OVOK-4k validation set.
Without these controls, early metric gains plateau quickly and the run stagnates, whereas the scheduled run continues to reduce \(K\), \(\mathbf{T}_{c\leftarrow b}\), and joint errors.

Concretely, the uncontrolled run never rediscovers the staged workflow of Sec.~\ref{sec:exp-rsi}.
In the first three iterations it treats RoboPEPP~\cite{goswami2025robopepp} fine-tuning and AnyCalib~\cite{tirado2025anycalib} inference tweaks as a single accuracy shortcut, rather than separating intrinsics from articulated geometry.
In iterations 4--6 it then tries to prune computation and raise FPS before geometric accuracy has stabilized, so efficiency-oriented edits fail verification.
In iterations 6--8 it attacks focal-depth ambiguity by alternating refinements of \(K\), \(\mathbf{T}_{c\leftarrow b}\), and joint angles from a weak initialization,
which further aggravates the underdetermined coupling among the three factors.
Later iterations (9--14) attach temporal-consistency modules. Accuracy gains remain small while latency rises sharply, so the verifier rejects them.
Across the trajectory, the agent never consolidates a planned, recursive self-improving search.

These behaviors highlight two failure modes of unconstrained LLM search~\cite{yuksekgonul2025textgrad,huang2026textresnet,xia2026rrsi,wu2026modularrsi,nguyen2026where,zhu2026fault,cai2026reliable}:
(i)~mis-attribution: the agent fails to recover the core bottleneck behind the verification report; for example, when mid-run geometric error remains large, it pursues efficiency edits instead of improving estimation quality; and
(ii)~schedule failure: macroscopically, the search lacks a physics-grounded plan such as \(K\!\rightarrow\!(\mathbf{T}_{c\leftarrow b},q)\!\rightarrow\!\mathrm{FPS}\) with stage-wise focus; at a finer grain, failure-driven local fixes yield overly narrow hypotheses that consume the iteration budget without resolving the coupled estimation problem.
The human-authored failure-mode vocabulary constrains attribution, while the staged schedule enforces edit planning and granularity.

%% file: sec/5_conclusion.tex
\section{Conclusion}
\label{sec:conclusion}

We presented OVOK, a recursive self-improving framework for video-based articulated robot state estimation.
Under a human-authored protocol, an initializer builds OVOK-4k to close the metric feedback loop; from a light seed, a proposer--executor--verifier cycle then jointly evolves geometry and perception tools and their workflows for a long horizon.
The retained workflow is frozen at inference and achieves state-of-the-art metric accuracy and robustness on DREAM-V and RoboKeyGen-V.
OVOK serves both as a practical path to metric robot state from video and as a template for long-horizon self-improvement in other geometric perception problems under coupled physical constraints.

%% file: sec/6_llm_usage.tex
\section*{LLM Usage Statement}

We used ChatGPT~(OpenAI) and DeepSeek as assistive tools throughout manuscript preparation, including drafting and restructuring text, refining technical descriptions, improving language, related work, and LaTeX layout, as well as for reviewer-style feedback on the manuscript.
In the research, multimodal foundation models (GPT-6.1 Sol, GPT-6 Astra, and Claude Opus~5.5) were used as OVOK agent backbones and as zero-shot baselines~(Sec.~\ref{sec:exp}).
LLM-based assistants also helped refine author-produced code, experimental analysis, figures, and synthetic-data protocols at an assistive level.
The authors take full responsibility for the methodology, experimental results, interpretations, and final manuscript.

%% file: sec/A_appendix.tex
\clearpage
\setcounter{page}{1}
\maketitlesupplementary
\appendix
\makeatletter
\def\ps@appsuppfirst{%
  \let\@oddhead\@empty
  \let\@evenhead\@empty
  \def\@oddfoot{%
    \reset@font
    \rlap{%
      \parbox[b]{\columnwidth}{\raggedright\footnotesize
        All the code and dataset are available by Dec.~2026, at:\\
        \url{https://github.com/dccc2025/ovok_official}.}}%
    \hfil\thepage\hfil
  }%
  \let\@evenfoot\@oddfoot
}
\thispagestyle{appsuppfirst}
\makeatother

\noindent\textbf{Contents.}
\begin{itemize}
  \item \ref{app:data}~~Initialization
  \item \ref{app:evolution}~~Evolution
  \begin{itemize}
    \item \ref{app:human-init}~~Human initialization
    \item \ref{app:obj-skills}~~Objective and skills
    \item \ref{app:proposer}~~Proposer setting
    \item \ref{app:verifier}~~Verifier setting
  \end{itemize}
  \item \ref{app:experiments}~~Experiments
  \begin{itemize}
    \item \ref{app:exp-setup}~~Experimental setup
    \item \ref{app:limitations}~~Limitations and future work
  \end{itemize}
\end{itemize}

\section{Initialization}
\label{app:data}

Initialization builds the data substrate used later for tool revision and workflow scoring.
Unlike prior task-specific datasets~\cite{lee2020dream,tian2024robokeygen,liu2026robotag}, OVOK-4k stresses harder observations:
varying intrinsics \(K\), diverse viewpoints and distances, self-occlusion, a small on-screen footprint, truncation, and complex lighting across clips.
Table~\ref{tab:synth-compare} compares OVOK-4k with the two synthetic corpora that dominate prior Franka RGB estimators:
DREAM-synth~\cite{lee2020dream} (also the usual training source of RoboPose, HoRoPose, and RoboPEPP~\cite{labbe2021robopose,ban2024horopose,goswami2025robopepp})
and RoboKeyGen's SimRGBD-Franka~\cite{tian2024robokeygen}.

\begin{table*}[t]
\centering
\footnotesize
\caption{Synthetic training sets for Franka RGB robot-state estimation.}
\label{tab:synth-compare}
\renewcommand{\arraystretch}{1.22}
\setlength{\tabcolsep}{6pt}
\begin{tabularx}{\textwidth}{@{}>{\raggedright\arraybackslash}p{2.35cm} *{3}{>{\centering\arraybackslash}X}@{}}
\toprule
 & DREAM-synth~\cite{lee2020dream}
 & SimRGBD-Franka~\cite{tian2024robokeygen}
 & OVOK-4k \\
\midrule
Scale
  & ${\sim}100$k images
  & $4$k $\times$ $3$ frames
  & $4$k $\times$ $75$ frames \\
Scene
  & randomized 3D scene
  & Blender scenes
  & RoboCasa kitchens \\
Temporal unit
  & individual samples
  & $3$-frame clips
  & $75$-frame clips ($25$\,Hz) \\
Camera distance
  & $0.75$--$1.20$\,m
  & unspecified ($1.29$--$1.82$\,m at test)
  & $0.85$--$2.55$\,m \\
Intrinsics \(K\)
  & known
  & known
  & unknown; \(f_x{=}f_y\in[298,625]\), \(c_x\in[300,340]\), \(c_y\in[160,280]\) \\
Motion
  & independent joint samples
  & simulated motion
  & gated rollouts (Tab.~\ref{tab:gates}) \\
Supervision
  & 2D keypoints
  & pose, joints, mask, depth
  & RGB + \((q,K,\mathbf{T}_{c\leftarrow b})\) \\
\bottomrule
\end{tabularx}
\end{table*}

Under the human-authored protocol in Fig.~\ref{fig:init-prompt}, a foundation-model initializer instantiates a RoboCasa~\cite{nasiriany2026robocasa365} simulation-and-rendering pipeline (50 layouts \(\times\) 50 styles) and emits paired RGB / simulator-state clips of a Franka Panda whose base is fixed in the world for the clip.
Each accepted clip is \(640{\times}480\) uint8 RGB at \(25\,\mathrm{Hz}\) for 75 frames; within a clip, \(K\) and \(\mathbf{T}_{c\leftarrow b}\) are constant, while both vary across clips.
Train (3{,}600) and valid (400) use disjoint room-layout IDs; a clip enters OVOK-4k only if it meets the numerical admission gates (see Tab.~\ref{tab:gates}).

\begin{table}[H]
\centering
\footnotesize
\caption{OVOK-4k admission gates.}
\label{tab:gates}
\setlength{\tabcolsep}{4pt}
\renewcommand{\arraystretch}{1.15}
\begin{tabular}{@{}l@{\hspace{6pt}}>{\raggedright\arraybackslash}p{0.68\columnwidth}@{}}
\toprule
Group & Gate \\
\midrule
Visibility &
robot mask $\geq 2\%$ of $640{\times}480$ (every frame) \\
& off-screen vertices $\leq 40\%$ \\
& robot occlusion $\leq 15\%$; base occlusion $\leq 0.5\%$ \\
& clearance $\geq 0.15$\,m (scene) / $0.20$\,m (robot) \\
& distance $0.85$--$2.55$\,m; base partly in view \\
\midrule
Physical &
FK $\leq 10^{-4}$\,m; re-projection $\leq 0.1$\,px \\
& penetration $=0$; dynamics replay $\|\Delta q\|<10^{-5}$ \\
& base/camera static within clip \\
& $f_x{=}f_y\in[298,625]$; $c_x\in[300,340]$, $c_y\in[160,280]$ \\
\midrule
Motion &
$\geq 3$ joints with range $>0.08$\,rad and hand $\geq 0.08$\,m \\
& or wrist-family orientation change $\geq 0.35$\,rad \\
\midrule
Diversity &
$\geq 40$ train / $12$ valid clips per family \\
& camera-grid occupancy $\geq 50\%$ \\
& $\leq 1$ accepted window per rollout \\
\bottomrule
\end{tabular}
\end{table}

\begin{figure*}[t]
  \centering
  \noindent{\color{black!30}\hrule height 0.5pt}
\begin{lstlisting}[
  style=ovokprompt,
  basicstyle=\scriptsize\ttfamily\color{black!88},
  frame=none,
  framesep=0pt,
  backgroundcolor=\color{black!3},
  xleftmargin=4pt,
  xrightmargin=4pt,
  aboveskip=4pt,
  belowskip=4pt
  ]
# Initialization

## Role
You are the Initializer for OVOK: a producer that builds a simulation-and-rendering pipeline and emits paired RGB / simulator-state videos for articulated robot state estimation.

## Goal
Construct **OVOK-4k** under a human-authored data contract:
- Train : 3,600 clips for tool revision
- Valid : 400 clips for workflow scoring
- Train and valid use **disjoint** RoboCasa room-layout IDs

## Scene prior
- Simulator / assets: RoboCasa (50 layouts x 50 styles)
- Robot: Franka Panda URDF, articulated gripper; base fixed in the world per clip
- Placement: random placed on the floor; physically valid contacts and no penetration
- Motion: smooth and normal trajectories spanning diverse motion families
- Camera: one world-fixed pinhole camera per clip; randomized viewpoint across clips
- Intrinsics `K` and camera-to-base pose `T_{c<-b}` are constant within a clip, but vary across clips

## Observation Coverage
Prefer observations that is hard to predict:
- Wide viewpoint diversity (azimuth / elevation / distance / roll)
- Challenge conditions, such as self-occlusion, truncation, and small robot footprint (still above visibility gates)
- Diverse lighting, distracting objects placed on the scene

## Output Pair contract
For video in pairs, each accepted clip is with:
- Resolution: `640 x 480`, `uint8`
- Length: `25 Hz`, `75` frames (`0.00-2.96 s`)
- Paired GT per frame: joints, 2D/3D keypoints, `K`, `T_{camera<-base}`,
  `T_{world<-camera}`, gripper opening, robot mask
For annoation in pairs, per-frame GT is written to `gt.json` under `clips[].frames[]`, including at least these terms: joint angles (unit: degree),  2D/3D keypoints, `K`, `T_{camera<-base}`, gripper opening, robot mask. They follow in form:
```json
{
  "joint_angles_rad": [0.0, 0.0, 0.0, 0.0, 0.0, 0.0, 0.0],
  "keypoints_2d": {
    "panda_link0": [u, v],
    "panda_link2": [u, v],
    "panda_link3": [u, v],
    "panda_link4": [u, v],
    "panda_link6": [u, v],
    "panda_link7": [u, v],
    "panda_hand": [u, v]
  },
  "keypoints_3d": {
    "panda_link0": [x, y, z],
    "panda_link2": [x, y, z],
    "panda_link3": [x, y, z],
    "panda_link4": [x, y, z],
    "panda_link6": [x, y, z],
    "panda_link7": [x, y, z],
    "panda_hand": [x, y, z]
  },
  "K": [
    [fx, 0.0, cx],
    [0.0, fy, cy],
    [0.0, 0.0, 1.0]
  ],
  "T_camera_from_base": [
    [r00, r01, r02, tx],
    [r10, r11, r12, ty],
    [r20, r21, r22, tz],
    [0.0, 0.0, 0.0, 1.0]
  ],
  "gripper_joint_positions_m": [left_m, right_m],
  "robot_mask_rle": {
    "size": [480, 640],
    "order": "C",
    "counts": [run0, run1, "..."]
  }
}
```

## Execution Procedure 
1. Sample layout, style, seeds, motion family, and camera cell
2. Instantiate the scene; place the Panda; roll out physics
3. Render RGB; dump simulator state at each exposure
4. Run the independent data gates (see paper Table for thresholds)
5. Accept only if all gates pass; otherwise reject and resample

\end{lstlisting}
  \noindent{\color{black!30}\hrule height 0.5pt}
  \caption{Human-authored prompts at the Initialization stage.}
  \label{fig:init-prompt}
\end{figure*}

\section{Evolution}
\label{app:evolution}

\subsection{Human initialization}
\label{app:human-init}
Evolution starts from a light human-authored seed
$(\mathcal{U}_0,\mathcal{S}_0,\mathcal{W}_0)$.
The seed remains frozen for the whole run, which keeps the objective from drifting as skills and tools evolve.
It orients the proposer--executor--verifier loop, but it neither interferes with exploration nor prescribes the retained solver.

\paragraph{Initialized skills.}
Seed skills let the proposer understand the task before any verification evidence exists.
The \emph{mathematical problem definition} (Fig.~\ref{fig:seed-problem-def}) states the estimation objective in compact geometric and kinematic form.
The \emph{background and motivation} (Fig.~\ref{fig:seed-background}) summarizes prior work in structural language and records the structural prior that articulated state estimation admits a modular decomposition into subproblems.
Together they specify \emph{what} to estimate and a prior for \emph{how} to compose a workflow.

\paragraph{Initialized tools.}
Initialization also seeds a compact tool set \(\mathcal{U}_0\) in three banks:
\emph{mutable tools}, \emph{frozen pretrained models}, and
\emph{foundation-model vision}~\cite{li2026roboharness,glantz2026seeing,han2025tiger,shi2026pluginrsi}.
Mutable tools are editable geometric and systems primitives: calibration, PnP, joint optimization, rendering, segmentation, tracking, and testing.
Frozen models supply perception and task-specific predictors that the workflow may call:
visual backbones Depth Anything~3~\cite{lin2025da3}, DINOv3~\cite{simeoni2025dinov3}, and V-JEPA~2~\cite{bardes2024vjepa},
together with robot estimators RoboPose~\cite{labbe2021robopose}, HoRoPose~\cite{ban2024horopose}, and RoboPEPP~\cite{goswami2025robopepp}.
These weights remain frozen; evolution may attach adapters or train small task-specific heads on the training split, with at most \(25\)M trainable parameters~\cite{bousmalis2023robocat,kulikov2026autodata}.
Foundation-model vision exposes a lightweight ask/answer interface for visual inspection or coarse zero-shot initialization.

\paragraph{Initialized workflow.}
The seed workflow \(\mathcal{W}_0\) is minimal (Algorithm~\ref{alg:seed-workflow} in the main paper).
AnyCalib~\cite{tirado2025anycalib} estimates intrinsics from the first frame; HoRoPose~\cite{ban2024horopose}
then recovers a per-frame pose and joint angles under the fixed \(\hat K\), after which
\(\hat{\mathbf{T}}_{c\leftarrow b}=\frac{1}{N}\sum_{t=1}^{N}\hat{\mathbf{T}}_{c\leftarrow b,t}\).
This workflow gives evolution a runnable end-to-end baseline and a measurable
starting point for \(J_\tau\).

\begin{figure*}[t]
  \centering
  \noindent{\color{black!30}\hrule height 0.5pt}
\begin{lstlisting}[
  style=ovokprompt,
  basicstyle=\scriptsize\ttfamily\color{black!88},
  frame=none,
  framesep=0pt,
  backgroundcolor=\color{black!3},
  xleftmargin=4pt,
  xrightmargin=4pt,
  aboveskip=4pt,
  belowskip=4pt
  ]
# Problem Definition

Video-based articulated robot pose and joint-angle estimation recovers a robot's
camera-relative pose and joint configuration over time from an image sequence.

## Significance and Usage

Video-based articulated robot pose and joint-angle estimation provides a geometric
description of the robot's motion in its surroundings, supporting vision-based
manipulation and closed-loop control, physically grounded 4D scene reconstruction,
and fine-grained motion planning.

## State and Observation

Let M be a known kinematic model of a robot: a tree of links, n revolute joints with
limits Q subset R^n, and a body frame on each link. A configuration q in Q determines,
by forward kinematics, a rigid pose FK_j(q) in SE(3) of every link j in the robot-base frame.

An RGB camera with intrinsics K records a video I_{1:N} = (I_t)_{t=1}^{N}, where
I_t in R^{H x W x 3}. At frame t, the robot state is (T_{c<-b}, q_t), where
T_{c<-b} in SE(3) maps the robot-base frame to the camera frame and q_t specifies
the joint configuration.

For kinematic keypoints J defined at selected link origins, the image projections are

    p_{t,j} = pi( K, T_{c<-b}, FK_j(q_t) ) in R^2,    j in J.

Conditioned on K and q_t, forward kinematics supplies 3D keypoints whose image
correspondences permit pose recovery through PnP under non-degenerate geometry.

## Task Definition

Given I_{1:N} and M, we estimate hat{K}, the camera-relative pose hat{T}_{c<-b},
and joint angles hat{q}_t in Q per frame.
We instantiate this on Franka Emika Panda, taking J as the seven link-origin markers.
\end{lstlisting}
  \noindent{\color{black!30}\hrule height 0.5pt}
  \caption{Human-initialized skills at the Evolution stage: mathematical problem definition.}
  \label{fig:seed-problem-def}
\end{figure*}

\begin{figure*}[t]
  \centering
  \noindent{\color{black!30}\hrule height 0.5pt}
\begin{lstlisting}[
  style=ovokprompt,
  basicstyle=\scriptsize\ttfamily\color{black!88},
  frame=none,
  framesep=0pt,
  backgroundcolor=\color{black!3},
  xleftmargin=4pt,
  xrightmargin=4pt,
  aboveskip=4pt,
  belowskip=4pt,
  firstline=1,
  lastline=25
  ]
# Background and Motivation

In this file, the author will hand-write the current related works, core opinions for the agent on how to find the proper way for evolvement.

## Related Works

Robot pose and joint-angle estimation primarily uses render-and-compare refinement or learned state prediction.
RoboPose iteratively matches rendered and observed images to refine pose and joint angles.
RoboKeyGen generates 3D keypoints from 2D detections via diffusion, while RoboPEPP learns keypoint and joint-angle predictors through masked-joint pretraining and recovers pose with PnP.
HoRoPose and RoboTAG incorporate depth predicted from RGB into pose and joint-angle estimation.
When joint angles are available, DREAM, CtRNet, SGTA-Pose, and CtRNet-X instead focus on camera-to-robot pose.
FEEPE uses kinematic motion priors and measured depth to refine end-effector poses, while SPDH estimates 3D joint locations from depth.
These additional inputs simplify estimation but restrict deployment without robot-state access or depth sensing.

Citations: `paper_list.md`. PDFs are not copied into the seed library.

These methods can be classified into 3 groups: render-and-compare, learning-based state prediction, and foundational model zero-shot prediction.

Render-and-compare mainly refers to the RoboPose, it use refiner CNN to predict the increment of joint angles, and camera-to-robot positions, which is valuable for us to especially use it for the overall refinement or partial joint refinement, and it is also the key way to guide a relatively accurate predictions to the more accurate one. However, it does not have good performance when small pixels occupation of the robot in the image, also it needs the full-robot vision, when it is partially visible or in self-occlusion, its performance degrades dramatically. So, if agent designs tools, be careful about this condition.

Learning-based state prediction directly predicts the joint angles and 2d pose, 2d pose is argmax-selected from the image. Here, the author mainly takes the HoRoPose, RoboPEPP, RoboKeyGen as examples. From the expertized opinion, it is useful, simple, direct and high-efficient way. However, though strong ability to prediction under the corner cases, but it struggles on achieving accurate and optimal estimation. Also, current methods are not fine and complete, for example, that for the common insights, the output should combine with the confidence, if it's high confidence, the solver will believe the results, but if with low confidence, the specific keypoint and joint estimation should be solved by other methods. So, if agent tries to borrow the ideas of current methods, u should make it from coarse version to fine, design proper methods to achieve high-accurate estimations.

Foundational model zero-shot prediction means that for a video, the agent will sample frames and predict the results based on their strong visual pretrained priors. Compared with render-and-compare and learning-based state prediction methods, it is less accurate due to less 'training', context, reliable tools to invoke, also lacks the mature and stable playbook-structured workflow (When a new conversation with the foundational model, they will repeat thinking the same thing, and solve the task blindly). That's what u need to do in the iterations. U should design and improve the context (mainly refered skills), tools and workflows to help foundational model at the initial stage quickly understand the domain knowledge, and predict the accurate robot pose and joint angles estimations in real-time way.
\end{lstlisting}
  \noindent{\color{black!30}\hrule height 0.5pt}
  \caption{Human-initialized skills at the Evolution stage: background and motivation.}
  \label{fig:seed-background}
\end{figure*}

\begin{figure*}[t]
  \ContinuedFloat
  \centering
  \noindent{\color{black!30}\hrule height 0.5pt}
\begin{lstlisting}[
  style=ovokprompt,
  basicstyle=\scriptsize\ttfamily\color{black!88},
  frame=none,
  framesep=0pt,
  backgroundcolor=\color{black!3},
  xleftmargin=4pt,
  xrightmargin=4pt,
  aboveskip=4pt,
  belowskip=4pt,
  firstline=27
  ]
## Expertized Analysis, Motivation

Based on the task specifications, the only inputs are videos and robot urdf (3d mesh), u need to real-time predict the 2d-3d correspondence, joint angles accurately.

Based on the previous works, like RoboPose, HoRoPose, RoboPEPP, RoboKeyGen, we impose the structural prior that `articulated state estimation admits a modular decomposition into subproblems`.

Current methods generally use a modular, serial pipeline. Across RoboPose, HoRoPose, RoboPEPP, and RoboKeyGen, keypoint localization and camera-relative pose recovery recur as separable subproblems. The task decomposes into camera calibration, 2D/3D keypoint localization, joint prediction, camera-to-base pose recovery, and temporal refinement; each of these can be decomposed in the same modular way.

Deployment, however, requires adaptive depth: well-observed clips need only a short route to accurate estimates, while occluded or ambiguous ones need heavier inference. Compose those modules into a branchable workflow rather than one fixed serial pipeline. A branchable workflow covers more corner cases than one serial pipeline, attributes failures to the responsible unit so fixes can be reused, and keeps each revision cheaper than retraining an end-to-end model. The modular design is efficient and extensible, and adapts to new observations through incremental updates to individual modules.

For example, take a short serial route on easy clips; on occlusion, a small image footprint, or an unstable `panda_link0`, branch to heavier tools only where the cheap path fails -- a render-and-compare refiner (RoboPose), Depth Anything V3, a foundation-model visual query, or an alternative estimator (HoRoPose, RoboPEPP) fused when a single module is unreliable.

So, u need to have the sophiscatedly overall and high-level thinking when the designing the tools, for example, whether it is reliable, how to achieve high-accurate and high-efficient workflow for the agent. The real-time requirements hint that the tools and workflows should not be complicated.

## Inverse process and workflow design

The inverse problem is mainly blocked by 2 key factors:

1. Subtle visual cues make it hard to locate keypoints and estimate joint angles accurately.
2. Unknown intrinsics and camera-to-robot pose couple perspective projection with articulation.

The first term is more complicated. Self-occlusion, partial visibility, and a small
image footprint (the robot occupies few pixels) commonly occur. Subtle base-orientation
cues (small visual difference between the robot base and the background) lead to wrong
estimation of `panda_link0`. Also, due to the small pixel footprint of the end-effector,
estimation of `hand` may incur large error. J3 and J5 likewise produce large angular
deviations because their visual cues are weak. The problems are not limited to the above.

For the second factor, typical failure modes include "translation trades focal length"
and "a wrong joint angle can hide behind a compensating T_{camera<-base}".
The problems are not limited to the above.

Reason from the physical model `p = pi(K, T_{c<-b}, FK(q))`, Panda joint limits,
and velocity smoothness. Prefer iterative algorithms that:

1. Lock a stable `K` early. Intrinsics do not change inside a clip and can be
   estimated from background environment texture.
2. Once `K` is judged adequate, refine `T_{c<-b}` and the joint estimators,
   which dominate final pose accuracy.
3. Enforce a temporal prior: within a short frame window, robot state
   (`T_{c<-b}`, `q`) varies smoothly and slowly; regularize or reject jumps
   that violate joint limits or velocity smoothness.
4. Add residual heads or small models only where the base chain is unstable
   (tiny robot, base heading, J7, occlusion).
5. For large-angle joint errors, (possibily J1, J3, J5, and J7), the workflow generate multiple
   candidate angles and a refiner to disambiguate.

\end{lstlisting}
  \noindent{\color{black!30}\hrule height 0.5pt}
  \caption{Human-initialized skills at the Evolution stage: background and motivation. (cont.)}
\end{figure*}

\begin{figure*}[p]
  \centering
  \includegraphics[width=\textwidth,height=0.90\textheight,keepaspectratio]{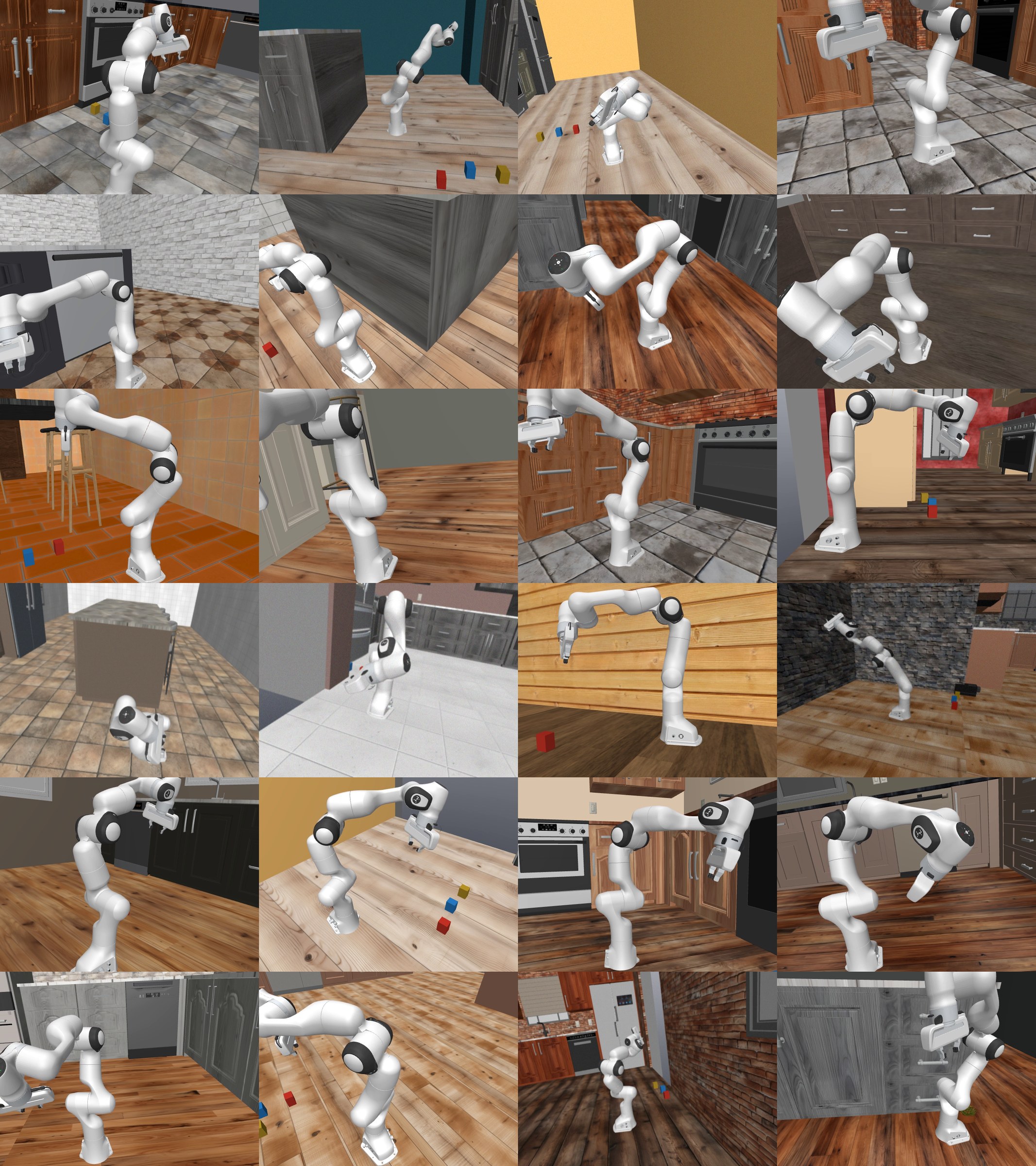}
  \caption{OVOK-4k sample frames across scene layouts, styles, and viewpoints.}
  \label{fig:ovok4k-samples}
\end{figure*}

\subsection{Objective and skills}
\label{app:obj-skills}

Eq.~\ref{eq:selection-objective} maximizes \(J_{\tau}\) over tools \(\mathcal{U}\) and workflow \(\mathcal{W}\) only.
Validation scores \(\textsc{Execute}(X_{\mathrm{v}};\mathcal{U},\mathcal{W})\), so \(J_{\tau}\) depends on the executable pair; skills \(\mathcal{S}\) are not an argument of the metric.
\(\mathcal{S}\) is retained as search memory: when a revision improves \(J_{\tau}\), accepted hypotheses are merged into \(\mathcal{S}\) and used by the next \textsc{Reason} stage, while inference still runs \(\textsc{Execute}(X;\mathcal{U}^{\star},\mathcal{W}^{\star})\).
Algorithm~\ref{alg:online-rsi} in the main paper makes this separation explicit.

\subsection{Proposer setting}
\label{app:proposer}

As described in the main paper, each evolution round runs three proposer stages:
\textsc{Reason}, \textsc{Revise}, and \textsc{Compose} (Algorithm~\ref{alg:proposer}).
\textsc{Reason} reads the previous verification report \(r_{i-1}\) to judge which edits
helped, and emits \(M\) hypotheses \(\{\delta_m\}_{m=1}^{M}\).
\textsc{Revise} updates tools under those hypotheses on the design split
\(\mathcal{D}_{\mathrm{tr}}'\):
it may rewrite mutable tools, fine-tune adapters, or train small task-specific models.
\textsc{Compose} then builds a candidate workflow \(\widetilde{\mathcal{W}}\) from
\(\{\delta_m\}\) and the revised tool set \(\widetilde{\mathcal{U}}\).

To keep RSI efficient, we do not send every hypothesis straight to full validation.
The 3{,}600-clip training set \(\mathcal{D}_{\mathrm{tr}}\) is partitioned once at the
start into a design split \(\mathcal{D}_{\mathrm{tr}}'\) (3{,}550 clips) and a fixed
dry-run probe \(\mathcal{D}_{\mathrm{dry}}\) (50 clips); the two sets are disjoint and fixed across iterations.
\textsc{Revise} fits on \(\mathcal{D}_{\mathrm{tr}}'\), and a hypothesis is rejected
early if it fails to improve on \(\mathcal{D}_{\mathrm{dry}}\)
(Algorithm~\ref{alg:proposer})\cite{agrawal2026gepa,zhang2025aflow,moll2026grasp}.
Only dry-run--supported candidates proceed to composition and the validation-time
executor--verifier cycle.

\begin{algorithm}[H]
\caption{Proposer with train dry-run rejection}
\label{alg:proposer}
\small
\begin{algorithmic}[1]
\Require task \(\tau\), \((\mathcal{U},\mathcal{S},\mathcal{W})\), report \(r\), train set \(\mathcal{D}_{\mathrm{tr}}'\) and \(\mathcal{D}_{\mathrm{dry}}\)
\Ensure revised tools \(\widetilde{\mathcal{U}}\), workflow \(\widetilde{\mathcal{W}}\), accepted hypotheses \(\mathcal{H}\)
\State \(J_{\mathrm{base}}\gets\textsc{Score}(\mathcal{W},\mathcal{U};\mathcal{D}_{\mathrm{dry}})\)
\State \(\{\delta_m\}_{m=1}^{M}\gets\textsc{Reason}(\tau,\mathcal{S},\mathcal{U},\mathcal{W},r,M)\)
\State \(\mathcal{H}\gets\varnothing\),\; \(\widetilde{\mathcal{U}}\gets\mathcal{U}\)
\For{\(m=1,\ldots,M\)}
    \State \(\mathcal{U}^{(m)}\gets\textsc{Revise}(\tau,\delta_m,\widetilde{\mathcal{U}},\mathcal{D}_{\mathrm{tr}}')\)
    \State \(\mathcal{W}^{(m)}\gets\textsc{Compose}(\tau,\delta_m,\mathcal{U}^{(m)},\mathcal{W})\)
    \State \(J^{(m)}\gets\textsc{Score}(\mathcal{W}^{(m)},\mathcal{U}^{(m)};\mathcal{D}_{\mathrm{dry}})\)
    \If{\(J^{(m)}>J_{\mathrm{base}}\)}
        \State \(\mathcal{H}\gets\mathcal{H}\cup\{\delta_m\}\)
        \State \(\widetilde{\mathcal{U}}\gets\mathcal{U}^{(m)}\),\; \(J_{\mathrm{base}}\gets J^{(m)}\)
    \EndIf
\EndFor
\State \(\widetilde{\mathcal{W}}\gets\textsc{Compose}(\tau,\mathcal{H},\widetilde{\mathcal{U}},\mathcal{W})\)
\State \Return \(\widetilde{\mathcal{U}}\), \(\widetilde{\mathcal{W}}\), \(\mathcal{H}\)
\end{algorithmic}
\end{algorithm}

\paragraph{Multi-term acceptance.}
The comparisons \(J_i>J^{\star}\) and \(J^{(m)}>J_{\mathrm{base}}\) in
Algorithms~\ref{alg:online-rsi} and~\ref{alg:proposer} decide whether a hypothesis
is retained.
A single scalar can stall early search or hide compensatory errors among \(K\), \(\mathbf{T}_{c\leftarrow b}\), and \(q\).
Guided by the kinematic image model \(\mathbf{p}=\pi(K,\mathbf{T}_{c\leftarrow b},\mathrm{FK}(q))\), we keep an unnormalized four-term vector
\begin{equation}
\mathbf{J}=(J_1,J_2,J_3,J_4),
\end{equation}
where \(J_1\) is the relative error of \(K\), \(J_2\) the MAE of \(\mathbf{T}_{c\leftarrow b}\), and \(J_3\) the MAE of the joint angles (all lower better), while \(J_4\) is serial end-to-end FPS (higher better).
Relative improvement and regression follow these directions.
A hypothesis is marked \textsc{supported} only if both hold:
(i)~at least one term improves by more than \(20\%\), or at least two terms improve by more than \(10\%\), while every remaining term regresses by less than \(2\%\);
(ii)~serial FPS stays at least \(0.3\).

\paragraph{Edit schedule.}
The gate above is fixed for the whole run; what \textsc{Reason} should attempt next is not.
A human-authored schedule (Fig.~\ref{fig:edit-schedule}) advises edit priority without a rigid iteration quota.
First lock \(K\), which can be estimated from background texture and is a prerequisite for metric recovery.
Once \(K\) is judged adequate, target \(\mathbf{T}_{c\leftarrow b}\) and the joint estimators, which dominate final pose accuracy.
Afterwards raise FPS only when accuracy does not regress.
\textsc{Reason} reads this schedule together with the previous report \(r_{i-1}\) when raising hypotheses.

\begin{figure*}[t]
  \centering
  \noindent{\color{black!30}\hrule height 0.5pt}
\begin{lstlisting}[
  style=ovokprompt,
  basicstyle=\scriptsize\ttfamily\color{black!88},
  frame=none,
  framesep=0pt,
  backgroundcolor=\color{black!3},
  xleftmargin=4pt,
  xrightmargin=4pt,
  aboveskip=4pt,
  belowskip=4pt
  ]
# Edit Schedule

## Role
You advise the Proposer on edit priority. This schedule is advisory: there is no rigid iteration quota.

## Macroscopic order
Reason from the physical model `p = pi(K, T_{c<-b}, FK(q))`. Prefer this order:

1. Lock camera intrinsics `K` first. Intrinsics do not change inside a clip and can be estimated from background environment texture. `K` is a prerequisite for metric recovery.
2. Once `K` is judged adequate, refine `T_{c<-b}` and the joint estimators, which dominate final pose accuracy.
3. After these geometric foundations meet the acceptance criteria, raise serial FPS only when accuracy does not regress.
\end{lstlisting}
  \noindent{\color{black!30}\hrule height 0.5pt}
  \caption{Human-authored edit schedule used by the proposer during evolution.}
  \label{fig:edit-schedule}
\end{figure*}

\subsection{Verifier setting}
\label{app:verifier}

After the executor runs a composed workflow on \(\mathcal{D}_{\mathrm{v}}\), the verifier returns a score \(\mathbf{J}_i\) and a report \(r_i\) (Algorithm~\ref{alg:online-rsi}).
As in Fig.~\ref{fig:loop}, the report has three parts---quantitative metrics, qualitative diagnosis, and failure attribution---and is what steers the next proposer round.
This follows recent RSI practice in which verification, rather than a scalar training loss, guides subsequent revisions~\cite{agrawal2026gepa,ye2026mce,yuksekgonul2025textgrad,zhang2026ace,xu2026harness2,srikanth2026aide2,zhang2025dgm}.

Quantitatively, the verifier evaluates \(\mathbf{J}=(J_1,J_2,J_3,J_4)\) on the validation split, together with auxiliary 2D/3D pose statistics used only for diagnosis.
A revision is retained only if it meets the multi-term rule in Sec.~\ref{app:proposer}.
Qualitatively, it samples high-error clips and diagnoses their failure patterns.
Attribution then maps those patterns onto eight canonical modes and nowhere else:
inaccurate intrinsics, severe \(\mathbf{T}_{c\leftarrow b}\) bias, global joint-angle bias, difficult base-orientation (J1), difficult end-effector (J7), occlusion/partial identifiability, lighting, and low computational efficiency.

The next \textsc{Reason} stage reads \(r_i\) together with the edit schedule in Sec.~\ref{app:proposer}.
Thus metrics decide whether to keep a revision, while the report decides what to edit next.

\begin{figure*}[t]
  \centering
  \noindent{\color{black!30}\hrule height 0.5pt}
\begin{lstlisting}[
  style=ovokprompt,
  basicstyle=\scriptsize\ttfamily\color{black!88},
  frame=none,
  framesep=0pt,
  backgroundcolor=\color{black!3},
  xleftmargin=4pt,
  xrightmargin=4pt,
  aboveskip=4pt,
  belowskip=4pt
  ]
# Verifier

## Role
You are the Verifier in OVOK's proposer--executor--verifier loop.
After the Executor runs a sealed workflow on the validation split, audit the
run and write a report that steers the next Proposer round.

## Division of labor
- Host code computes metrics and GT-vs-Pred diagnostic figures.
- You synthesize the narrative: what failed, which J axes matter, what to try next.
- Host selection owns the champion pointer (`confirmed.json`). If there is no
  prior champion, the first complete run is confirmed even when this report
  recommends reject.

## Quantitative pass
1. Read `/ovok/metrics.json`, focusing on the raw objective
   **J = (J1, J2, J3, J4)** (unnormalized):
   - J1: relative error of camera intrinsics K (lower better)
   - J2: MAE of camera-to-base pose T_{c<-b} (metres; lower better)
   - J3: mean MAE of the 7 joint angles (degrees; lower better)
   - J4: serial end-to-end FPS (higher better)
2. Apply the same multi-term gate as the host .
   Mark the run **supported** if either:
   - (i) at least three terms improve by more than 5% vs the incumbent, or
   - (ii) at least one term improves by more than 5% and every other term
     regresses by less than 1%.
   The first champion may still be accepted with a weak J when coverage=1.
3. Write `verifier_decision.json` to match this gate
   (`require_verifier_accept: true` on the host).

## Qualitative pass
4. Inspect diagnostic middle-frame error figures (white = GT, blue = Pred).
   Describe failures visually and geometrically (e.g., unstable intrinsics,
   distal-joint error under occlusion, throughput collapse from an
   over-composed workflow).
5. Write `review_reports.md` with:
   Quantitative Analysis (include raw J),
   Qualitative Analysis,
   Failure Modes,
   Weaknesses / Urgent Improvements (name stalled or regressed J axes and
   what to fix next),
   Next Iteration,
   Limitations.
6. Write `rsi_update.json` with phenomenon-level failure modes and one next
   experiment that targets the weakest J axis with a kinetics-aware redesign.
\end{lstlisting}
  \noindent{\color{black!30}\hrule height 0.5pt}
  \caption{Human-authored verifier prompt used during online evolution.}
  \label{fig:verifier-prompt}
\end{figure*}

\section{Experiments}
\label{app:experiments}

\subsection{Experimental setup}
\label{app:exp-setup}

All experiments run in a sandbox with 32 cores of an AMD EPYC 7742, one NVIDIA
A800 GPU ($80\,\mathrm{GB}$), $128\,\mathrm{GB}$ host memory, and a $1\,\mathrm{TB}$
storage limit (see Sec.~\ref{sec:exp-impl}).
We supplement the foundation-model experiments here.
Standalone baselines query the official APIs of Opus~5.5, GPT-6 Astra, and GPT-6.1 Sol at the medium reasoning-effort setting; each configuration is repeated for 10 independent runs.
The same APIs serve as OVOK backbones: we run 15 RSI iterations and evaluate the best retained workflow on DREAM-V and RoboKeyGen-V.
Unless otherwise noted, qualitative comparisons and ablations use GPT-6.1 Sol.

\paragraph{Sparse-sample foundation-model inference.}
Video VLMs are queried on sparsely sampled frames rather than at native frame rate~\cite{lin2024videollava}, and video pose estimators densify those keyframes by temporal propagation~\cite{bertasius2019posewarper}.
We follow the same protocol for a fair comparison: query the foundation model every 10 frames and hold each estimate until the next keyframe (Algorithm~\ref{alg:fm-sparse}).
Videos and keyframe calls are both serial; reported FPS uses pure processing milliseconds (API compute time).
Figures~\ref{fig:fm-prompt-dream} and~\ref{fig:fm-prompt-robokeygen} give the prompts used on DREAM-V and RoboKeyGen-V; each asks for seven URDF-link 2D keypoints and seven joint angles on the queried frame.

\begin{algorithm}[H]
\caption{Sparse-sample foundation-model inference}
\label{alg:fm-sparse}
\small
\begin{algorithmic}[1]
\Require videos \(\{I^{(v)}_{1:F_v}\}_{v=1}^{V}\), foundation model \(\mathcal{F}\), prompt \(\Pi\), robot model \(\mathcal{M}\)
\Ensure per-video predictions \(\{\widehat{Y}^{(v)}\}_{v=1}^{V}\), processing time \(T_{\mathrm{proc}}\)
\State \(T_{\mathrm{proc}}\gets 0\)
\For{\(v=1,\ldots,V\)}
    \For{\(t=1,11,\ldots,\le F_v\)}
        \State \(t_0\gets\textsc{Clock}()\)
        \State \(y_t\gets\mathcal{F}.\textsc{Ask}(\Pi,I^{(v)}_{t};\mathcal{M})\)
        \State \(T_{\mathrm{proc}}\gets T_{\mathrm{proc}}+(\textsc{Clock}()-t_0)\)
    \EndFor
    \For{\(\tau=1,\ldots,F_v\)}
        \State \(t^{\star}\gets 1+10\left\lfloor(\tau-1)/10\right\rfloor\)
        \State \(\widehat{Y}^{(v)}_{\tau}\gets y_{t^{\star}}\)
    \EndFor
\EndFor
\State \Return \(\{\widehat{Y}^{(v)}\}_{v=1}^{V}\),\; \(\mathrm{FPS}=(\sum_{v}F_v)/T_{\mathrm{proc}}\)
\end{algorithmic}
\end{algorithm}

\begin{figure*}[t]
  \centering
  \noindent{\color{black!30}\hrule height 0.5pt}
\begin{lstlisting}[
  style=ovokprompt,
  basicstyle=\scriptsize\ttfamily\color{black!88},
  frame=none,
  framesep=0pt,
  backgroundcolor=\color{black!3},
  xleftmargin=4pt,
  xrightmargin=4pt,
  aboveskip=4pt,
  belowskip=4pt
  ]
# Foundation-model baseline (DREAM-V)

## Role
You are a robot-perception agent.
Estimate the Franka Emika Panda state from RGB video: the base and arm move,
and joint angles are unknown. Prefer precise numeric estimates over analysis.

## Dataset
DREAM-Video (DREAM-V): 60 real RGB clips, 640x480, 30 fps.
Cameras: Azure Kinect, Kinect 360, RealSense, Orb.

## Inputs
- RGB videos for this task (`data/videos/` or the uploaded `.mp4` files).
- `panda.urdf` as the kinematic source of truth.

## Targets (every frame of every video)
1. Seven 2D keypoints in pixels (URDF link origins; not mesh centroids or joint
   vertices), in order:
   `panda_link0`, `panda_link2`, `panda_link3`, `panda_link4`,
   `panda_link6`, `panda_link7`, `panda_hand`.
2. Seven revolute joint angles in degrees, URDF order
   `panda_joint1` ... `panda_joint7`.

Keypoints and joints are not one-to-one:
- `panda_link0` is the base origin.
- `panda_link2` is at the shoulder (joint1/joint2 origins coincide).
- `panda_link6` is at the wrist (joint5/joint6 origins coincide).
- `panda_hand` follows fixed joints beyond link7.

Image axes: origin at the top-left; x right, y down.

## Procedure
1. Read the output schema for `predictions.json`.
2. Read `panda.urdf`.
3. For each video and frame, estimate the seven keypoints and seven joint angles.

## Allowed
- Inspect the RGB videos (or sampled frames).
- Use `panda.urdf` and your own visual reasoning.
- Write only `predictions.json`.

## Forbidden
- Do not call, download, or wrap published robot-pose / joint networks
  (e.g., RoboPEPP, RoboPose, HoRoPose, CtRNet, DREAM-ResNet), their weights,
  scripts, or existing predictions.
- If a frame is hard, still return your best numbers.

## Output schema (`predictions.json` only)
{
  "task_id": "ovok/dream-v",
  "dataset": "DREAM-V",
  "model": "<model_name>",
  "clips": [
    {
      "video": "panda-3cam_azure_01.mp4",
      "frames": [
        {
          "frame_index": 0,
          "keypoints_2d": {
            "panda_link0": [x, y],
            "panda_link2": [x, y],
            "panda_link3": [x, y],
            "panda_link4": [x, y],
            "panda_link6": [x, y],
            "panda_link7": [x, y],
            "panda_hand": [x, y]
          },
          "joint_angles_deg": [j1, j2, j3, j4, j5, j6, j7]
        }
      ]
    }
  ]
}
- One clip per video; `video` is the basename only.
- `frame_index` starts at 0 and is consecutive.
- `joint_angles_deg` has seven degrees for `panda_joint1`..`panda_joint7`.
\end{lstlisting}
  \noindent{\color{black!30}\hrule height 0.5pt}
  \caption{Prompt for foundation-model baselines on DREAM-V.}
  \label{fig:fm-prompt-dream}
\end{figure*}

\begin{figure*}[t]
  \centering
  \noindent{\color{black!30}\hrule height 0.5pt}
\begin{lstlisting}[
  style=ovokprompt,
  basicstyle=\scriptsize\ttfamily\color{black!88},
  frame=none,
  framesep=0pt,
  backgroundcolor=\color{black!3},
  xleftmargin=4pt,
  xrightmargin=4pt,
  aboveskip=4pt,
  belowskip=4pt
  ]
# Foundation-model baseline (RoboKeyGen-V)

## Role
You are a robot-perception agent.
Estimate the Franka Emika Panda state from RGB video: the base and arm move,
and joint angles are unknown. Prefer precise numeric estimates over analysis.

## Dataset
RoboKeyGen-Video (RoboKeyGen-V): 24 real RGB clips, 640x360, 30 fps.
Cameras: RealSense D415, Azure Kinect.

## Inputs
- RGB videos for this task (`data/videos/` or the uploaded `.mp4` files).
- `panda.urdf` as the kinematic source of truth.

## Targets (every frame of every video)
1. Seven 2D keypoints in pixels (URDF link origins; not mesh centroids or joint
   vertices), in order:
   `panda_link0`, `panda_link2`, `panda_link3`, `panda_link4`,
   `panda_link6`, `panda_link7`, `panda_hand`.
2. Seven revolute joint angles in degrees, URDF order
   `panda_joint1` ... `panda_joint7`.

Keypoints and joints are not one-to-one:
- `panda_link0` is the base origin.
- `panda_link2` is at the shoulder (joint1/joint2 origins coincide).
- `panda_link6` is at the wrist (joint5/joint6 origins coincide).
- `panda_hand` follows fixed joints beyond link7.

Image axes: origin at the top-left; x right, y down.

## Procedure
1. Read the output schema for `predictions.json`.
2. Read `panda.urdf`.
3. For each video and frame, estimate the seven keypoints and seven joint angles.

## Allowed
- Inspect the RGB videos (or sampled frames).
- Use `panda.urdf` and your own visual reasoning.
- Write only `predictions.json`.

## Forbidden
- Do not call, download, or wrap published robot-pose / joint networks
  (e.g., RoboPEPP, RoboPose, HoRoPose, CtRNet, DREAM-ResNet, RoboKeyGen),
  their weights, scripts, or existing predictions.
- If a frame is hard, still return your best numbers.

## Output schema (`predictions.json` only)
{
  "task_id": "ovok/robokeygen-v",
  "dataset": "RoboKeyGen-V",
  "model": "<model_name>",
  "clips": [
    {
      "video": "1_D415_front_0_01.mp4",
      "frames": [
        {
          "frame_index": 0,
          "keypoints_2d": {
            "panda_link0": [x, y],
            "panda_link2": [x, y],
            "panda_link3": [x, y],
            "panda_link4": [x, y],
            "panda_link6": [x, y],
            "panda_link7": [x, y],
            "panda_hand": [x, y]
          },
          "joint_angles_deg": [j1, j2, j3, j4, j5, j6, j7]
        }
      ]
    }
  ]
}
- One clip per video; `video` is the basename only.
- `frame_index` starts at 0 and is consecutive.
- `joint_angles_deg` has seven degrees for `panda_joint1`..`panda_joint7`.
\end{lstlisting}
  \noindent{\color{black!30}\hrule height 0.5pt}
  \caption{Prompt for foundation-model baselines on RoboKeyGen-V.}
  \label{fig:fm-prompt-robokeygen}
\end{figure*}

\subsection{Limitations and future work}
\label{app:limitations}

\paragraph{Limitations.}
Human-authored priors: the frozen seed, modular evolution policy, initialized tool pool, edit schedule, and multi-term acceptance gate are not ablated, so we cannot claim they are optimal.
We also do not vary the verifier (metrics, qualitative report, or attribution), leaving its effect on the retained workflow unmeasured.
Reported runs are further confined to offline, clip-level monocular RGB of a single Panda; streaming accuracy and cross-embodiment transfer remain open.
Finally, OVOK still requires a human task specification and human-in-the-loop steering of a frontier model: it is verification-guided search around that prior, not fully autonomous recursive self-improvement.

\paragraph{Future work.}
A first direction is more accurate, real-time streaming video estimation: update camera and robot state as frames arrive, rather than after the clip is complete, at latency compatible with closed-loop control.
A second is to test the same self-evolving loop on more complex metric measurement~\cite{dai2026hadar,dai2026tex1500,chen2026t2exture,you2026frontierphysics,tang2025ticoss}.